\documentclass[lettersize,journal]{IEEEtran}
\usepackage{flushend}
\usepackage{amsmath,amsfonts}
\usepackage[linesnumbered,boxed,ruled]{algorithm2e}
\usepackage{tabularx}
\usepackage{array}
\usepackage{graphicx}
\usepackage{booktabs}
\usepackage{textcomp}
\usepackage{stfloats}
\usepackage{url}
\usepackage{verbatim}
\usepackage{graphicx}
\usepackage{cite}
\usepackage{color}
\usepackage{graphicx}
\usepackage{subfigure}
\usepackage{float}
\begin{document}

\title{
Wireless Power Transfer and Intent-Driven Network Optimization in AAVs-assisted IoT for 6G Sustainable Connectivity}

% Intent-driven Network Optimization in AAVs-assited IoT

\author{
	%Yue Hu, 
    Xiaoming He, \emph{Member, IEEE}, Gaofeng Wang, Huajun Cui, Rui Yuan, and Haitao Zhao, \emph{Senior Member, IEEE} %and Shahid Mumtaz, \emph{Senior Member, IEEE}

	\thanks{
		%Y. Hu is with the College of Computer Science and Software Engineering, Hohai University, Nanjing 210024, China (E-mail: huyue4900@163.com).
		%X. He is with the College of Internet of Things, Nanjing University of Posts and Telecommunications, Nanjing, China (E-mail: hexiaoming@njupt.edu.cn). 

        X. He, and G. Wang are with the College of Internet of Things, Nanjing University of Posts and Telecommunications, Nanjing, China (E-mail: hexiaoming@njupt.edu.cn, wang123456az@gmail.com).
		
		H. Cui and R. Yuan are with the Digital Intelligence Research Institute, PowerChina, Beijing Engineering Corporation Limited, Beijing, China (E-mail: yuanr@bjy.powerchina.cn, cuihuajun@bjy.powerchina.cn).

        %H. Cui is with Digital Intelligence Research Institute, PowerChina, Beijing Engineering Corporation Limited, Beijing, China (E-mail: cuihuajun@bjy.powerchina.cn).

        H. Zhao is with the College of Telecommunications and Information Engineering, Nanjing University of Posts and Telecommunications, Nanjing, China (E-mail: zhaoht@njupt.edu.cn).
        
		%S. Mumtaz is with the Department of Engineering, Nottingham Trent University, Nottingham, U.K. and also with the Department of Electronic Engineering, Kyung Hee University, Gyeonggi, South Korea (E-mail: dr.shahid.mumtaz@ieee.org).
	}

	\thanks{Corresponding Author: Haitao Zhao
}}



\maketitle

\begin{abstract}
Autonomous Aerial Vehicle (AAV)-assisted Internet of Things (IoT) represents a collaborative architecture in which AAV allocate resources over 6G links to jointly enhance user-intent interpretation and overall network performance. Owing to this mutual dependence, improvements in intent inference and policy decisions on one component reinforce the efficiency of others, making highly reliable intent prediction and low-latency action execution essential. Although numerous approaches can model intent relationships, they encounter severe obstacles when scaling to high-dimensional action sequences and managing intensive on-board computation.

We propose an Intent-Driven Framework for Autonomous Network Optimization comprising prediction and decision modules. First, implicit intent modeling is adopted to mitigate inaccuracies arising from ambiguous user expressions. For prediction, we introduce Hyperdimensional Transformer (HDT), which embeds data into a Hyperdimensional space via Hyperdimensional vector encoding and replaces standard matrix and attention operations with symbolic Hyperdimensional computations. For decision-making, where AAV must respond to user intent while planning trajectories, we design Double Actions based Multi-Agent Proximal Policy Optimization (DA-MAPPO). Building upon MAPPO, it samples actions through two independently parameterized networks and cascades the user-intent network into the trajectory network to maintain action dependencies.

We evaluate our framework on a real IoT action dataset with authentic wireless data. Experimental results demonstrate that HDT and DA-MAPPO achieve superior performance across diverse scenarios.
\end{abstract}


\section{Introduction}
As Autonomous Aerial Vehicle (AAV)-assisted Internet of Things (IoT) systems continue to expand \cite{wei2022uav} and traffic behaviors grow increasingly heterogeneous \cite{cheng2023ai}, conventional architectures based on static manually configured forwarding rules reveal fundamental scalability limitations. To overcome these inefficiencies, Intent-Based Networking (IBN) \cite{wei2020intent} has emerged as a paradigm capable of transforming abstract operational objectives into self-validating configurations, enabling continuous autonomous adaptation to evolving large-scale IoT environments. By abstracting policies and coordinating intent-driven control over 6G infrastructures, Intent-Based IoT (IB-IoT) \cite{cerroni2017intent} fosters tightly coupled interactions between network operations and user demand; under AAV-enhanced optimization, both user experience and infrastructure responsiveness can be further strengthened.

Within IB-IoT, user intent prediction and IBN optimization are central tasks that link intent semantics to executable policies. Leveraging intent-oriented coordination, these components translate high-level user goals into real-time control actions, refine intent policies at the edge with minimal latency, and dynamically optimize the network through AAV assistance.

Existing IB-IoT studies \cite{wu2020framework, attkan2022cyber, custodio2024comparing} primarily depend on deterministic policy engines or heuristic optimization, which typically model intent-to-configuration mappings through linear or predefined transformations. Recent advances introduce deep learning intent predictors and Deep Reinforcement Learning (DRL)-based optimizers to capture temporal intent evolution and non-linear performance dependencies. Despite these improvements, intent-driven control still encounters several unresolved challenges:
\begin{itemize}
	\item \textbf{Long-Context Intent Parsing with Edge Constraints:} Ultra-long interaction histories require accurate semantic retention under strict computation and energy limits, yet vanishing gradients and memory bottlenecks degrade temporal consistency and increase inference delay \cite{pawar2024and}.
	\item \textbf{Expansive Action-Space Optimization Traps:} High-dimensional double action spaces trigger combinatorial explosion, causing DRL agents to converge to inferior local solutions \cite{jiang2022efficient, banik2023continuous}.
	\item \textbf{Intent-Expression Noise and Domain Shift:} Non-rigorous or ambiguous user expressions introduce uncertain labels, misalign intent–policy mappings, and lead to severe performance drops when deployed beyond seen domains \cite{purohit2015intent}.
\end{itemize}

To address these issues, our work constrains intent modeling to the implicit regime by mining historical behavior patterns, thereby avoiding ambiguity and noise from explicit inputs. We then introduce the Hyperdimensional Transformer (HDT) for traffic prediction, which embeds raw data into a hyperdimensional vector space, replacing conventional floating-point matrix operations with symbolic vector computation—maintaining prediction fidelity while reducing computational and energy overhead. For large-scale continuous control in AAV-assisted IoT, we further propose \underline{D}ouble \underline{A}ctions based \underline{M}ulti-\underline{A}gent \underline{P}roximal \underline{P}olicy \underline{O}ptimization (DA-MAPPO), which samples actions via decoupled networks and constructs the final sequence through composition, mitigating exponential joint-action sampling complexity.

The main contributions of this paper are summarized as follows.
\begin{itemize}
	\item We establish an end-to-end IBN architecture centered on implicit intent modeling. To characterize user mobility and complex wireless conditions, we incorporate the Continuum Crowds model for trajectory representation and integrate large- and small-scale fading models to evaluate time-varying delay across each 2.4 GHz link.
	\item HDT introduces hyperdimensional vector encoding as a substitute for conventional embedding and transforms attention similarity operations into symbolic computations, significantly reducing floating-point operations and energy cost. DA-MAPPO employs a hybrid strategy combining independent parameterized sampling with cascaded coupling, supporting parallel exploration while preserving high-order dependencies and avoiding local minima in high-dimensional action spaces.
	\item We conduct performance evaluation on a real IoT behavioral dataset and implement a simulation environment utilizing measured 2.4 GHz channel parameters. Experimental results show that, under diverse channel and resource constraints, HDT and DA-MAPPO consistently surpass existing baselines in prediction accuracy and decision efficiency, demonstrating reliability and robustness.
\end{itemize}

The remainder of this paper is structured as follows. Section II reviews related work. Section III defines the system model. Section IV formulates the optimization problem. Section V details the proposed methodology. Section VI presents the experimental results. Section VII concludes this paper.

\section{Related Work}
% In this section, we focus on traffic prediction and large model's application to it.
In this section, we discuss the relevant work and research trends from three aspects, i.e., IBN, intent prediction, and network optimization.

\subsection{Intent-based Networking}
Research on IBN has seen significant advances, with scholars proposing innovative solutions across various dimensions. 

\textbf{Foundational Theories:} The journey of IBN research began with Pang \textit{et al.} \cite{pang2020survey}, who defined IBN as a self-driving network that automates application intents via decoupled network control logic and closed-loop orchestration. Subsequently, Zeydan \textit{et al.} \cite{zeydan2020recent} analyzed recent progress in IBN, focusing on network management and orchestration. These studies laid the theoretical foundation for future research.

\textbf{Architectural Designs:} Building on the theoretical foundation, Han \textit{et al.} \cite{han2016intent} proposed an intent-based virtual network management platform for SDN, with the aim of automating virtual network management based on tenant intents. Furthermore, Ouyang \textit{et al.} \cite{ouyang2022brief} presented a generic IBN architecture, defined intent refinement, and proposed a standard classification method, which provided structural support for practical applications.

\textbf{Practical Applications:} In terms of practical applications, Yang \textit{et al.} \cite{yang2023smart} introduced the motivations of IBN and proposed a novel management architecture named SMART. In a similar vein, Elkhatib \textit{et al.} \cite{elkhatib2017charting} proposed an IBN mechanism in which applications declare abstract intents for the network, thereby enhancing the fulfillment of application requirements and optimization of network resources.
Finally, Kiran \textit{et al.} \cite{kiran2018enabling} presented the iNDIRA tool, which interacts with SDN north-bound interfaces to enable IBN for scientific applications.

These contributions collectively push the frontiers of IBN, yet challenges such as user intent comprehension and system implementation remain open research questions that warrant further exploration.

\subsection{Intent Prediction}
The field of intent prediction has seen several innovative advancements. 

\textbf{Network Management and Recommendation Systems:} The field of intent prediction has seen significant advancements with contributions from various researchers. Collet \textit{et al.} \cite{collet2022lossleap} proposed LossLeaP, a novel IBN forecasting model that aims to minimize target management objectives. Similarly, Li \textit{et al.} \cite{li2019graph} introduced the Graph Intention Network (GIN), which addresses behavior sparsity and weak generalization in click-through rate prediction for sponsored search by leveraging a co-occurrence commodity graph and multi-layered graph diffusion. These approaches highlight the diverse applications of intent prediction in network management and recommendation systems.

\textbf{Transportation and Smart Homes:} In other domains, Zyner \textit{et al.} \cite{zyner2019naturalistic} developed a method for predicting driver intent at urban intersections. This method employs recurrent neural networks with a mixture density network output layer. Meanwhile, Xiao \textit{et al.} \cite{xiao2023know} presented SmartUDI, a novel approach for accurate User Device Interaction (UDI) prediction in smart home environments. These studies demonstrate the potential of intent prediction in transportation and smart home applications.

\textbf{Advertising and Multi-Stage Prediction:} Additionally, Guo \textit{et al.} \cite{guo2020deep} proposed DSPN, a two-stage deep prediction network that jointly models advertiser intent and satisfaction through advertiser action features and performance indicators. This work further expands the scope of intent prediction into advertising and multi-stage prediction scenarios.

Although these studies focus on different application scenarios, they collectively push the boundaries of intent prediction across various domains. 
However, in resource-constrained edge IoT, the challenge of accurately capturing and interpreting user intent while reducing inference time and computational resources in ultra-long contexts remains an open research question worthy of further investigation.

\subsection{Network Optimization}
The field of user intent prediction has witnessed remarkable progress through various innovative approaches. These methods can be categorized into different dimensions depending on their applications and techniques.

\textbf{Network Topology and Resource Allocation:} Li \textit{et al.} \cite{li2023network} proposed DRL-GS, a novel deep reinforcement learning algorithm designed for network topology optimization. This approach combines a verifier, graph neural network, and DRL agent to efficiently search large topology spaces, demonstrating superior performance in real-world case studies. Meanwhile, Fang \textit{et al.} \cite{fang2022drl} designed a DRL-driven intelligent optimization strategy for resource allocation in cloud-edge-end cooperation environments. They introduced a novel DRL policy that enhances content distribution by leveraging user history, cache capacity, link bandwidth, and topology structure.
 
\textbf{Routing Optimization and Computation Offloading:} In the realm of routing optimization, He \textit{et al.} \cite{he2023routing} presented MPDRL, a scheme that integrates graph neural networks with deep reinforcement learning. Furthermore, Wang \textit{et al.} \cite{wang2020utility} proposed a DRL-based method for joint optimization of radio and computation resources in network slicing with MEC. Additionally, Chen \textit{et al.} \cite{chen2021drl} proposed TADPG, a novel DRL agent for joint computation offloading and resource allocation in MEC. This agent employs temporal feature extraction and prioritized experience replay to enhance training convergence, outperforming state-of-the-art DRL agents in terms of task completion time and energy consumption.
 
\textbf{Edge-Cloud Networks and Optical Networks:} Ullah \textit{et al.} \cite{ullah2023optimizing} proposed the DDQNEC scheme, which utilizes the DDQN algorithm to optimize task offloading and resource allocation in edge-cloud networks. In another direction, Almasan \textit{et al.} \cite{almasan2022deep} integrated Graph Neural Networks (GNNs) into DRL agents, enabling generalization over unseen network topologies. Their DRL+GNN agent has shown outstanding performance in routing optimization for optical networks.
 
\textbf{Distributed Storage Resource Management:} Finally, Wang \textit{et al.} \cite{wang2021incorporating} proposed a distributed DRL-based storage resource management algorithm for SAGIN, further expanding the application scope of user intent prediction methods.

Collectively, these studies demonstrate the potential to adapt the network to user intent through DRL and GNN integration, but most edge nodes such as AAV still need to plan their own paths while adjusting the network, and there are still challenges in the selection of action sequences composed of multiple dimensions.

\subsection{Our Motivation}
Urbanization and IoT proliferation have driven IBN to couple deep semantic parsing with distributed intelligent decision-making, enhancing user engagement via intent–resource joint modeling and context-adaptive orchestration. Despite significant progress, two core challenges persist: (1) accurately capturing and interpreting user intent within ultra-long contexts while compressing inference latency and resource consumption, as current long-sequence models suffer information dilution and computational explosion; (2) multi-dimensional action-sequence selection remains inefficient due to the vast search space, with existing algorithms prone to local optima. 

To this end, we introduce a Hyperdimensional Transformer that replaces floating-point matrices with vector operations, sharply reducing complexity, and design DA-MAPPO, which retains action dependencies via cascaded networks to achieve optimal trade-offs in high-dimensional action spaces.

\begin{figure*}[t]
\centering 
\includegraphics[width=1\textwidth]{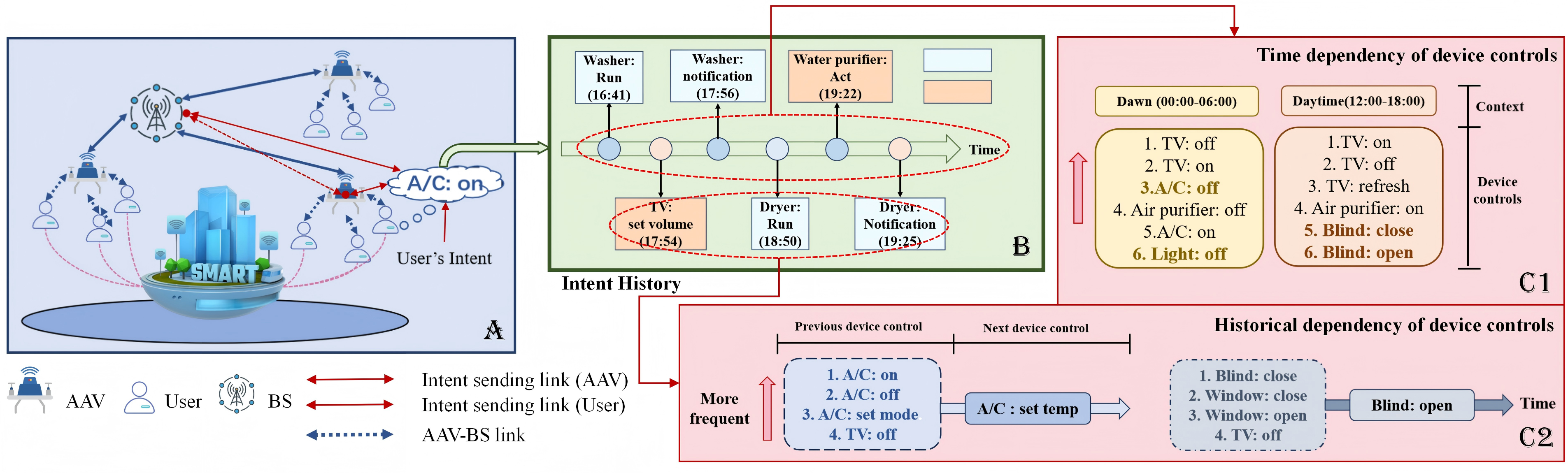}
\caption{The framework of Intent-based Networking.}
\label{network}
\end{figure*}

\section{System Model}\label{B4}
This section first establishes a comprehensive network model for AAV-assisted IoT. Next, we standardize the types of user intent conveyed within this network. Considering user mobility, we leverage an inherent human mobility model to guide the evolution of user locations. To faithfully capture the complexities of real-world communications, we consolidate the fading characteristics of the wireless channel and construct an AAV communication model.

\subsection{Intent-based Network Model}
We consider an IBN comprising one Base Station (BS), $N_v$ AAVs $\{V_1,V_2,\dots,V_i,\dots,V_{N_v}\}$, and $N_u$ users $\{U_1,U_2,\dots,U_j,\dots,U_{N_u}\}$ to assist the IoT.  
Within this IoT, each user at time $t$ triggers an intent $R_{U_j}(t)$ to interact with its smart devices; this intent is either served exclusively by one AAV as $R_{U_j}^{V_i}(t)$ or forwarded by an AAV to the BS as $R_{U_j}^{BS}(t)$.  
Every AAV is restricted to serve exactly $N_l$ users, so some intents may remain unscheduled; in such cases the user autonomously submits the intent to the BS, denoted by $R_{U_j}^{U_j}(t)$.

Fig.~\ref{network}(a) illustrates the IBN-assisted intent-response workflow.  
Users neither describe nor submit intents directly; instead, they authorize historical control information to the IBN, enabling it to infer latent demands.  
The historical control information comprises device states and usage time; the IBN extracts this information to deduce the user intent and proactively prepares the next action, thereby mitigating misinterpretation caused by non-expert user descriptions.

\subsection{User Intent Model}
This subsection details the IBN-centric user intent.  
Fig.~\ref{network}(b) presents the user-authorized historical information.  
When the user controls laundry-related devices (washer and dryer), their long, non-interactive cycles allow simultaneous control of other devices (TV and water purifier) during the same interval.  
Consequently, the action sequence $R_{U_j}(t)$ within one slot exhibits volatile intent.  
Hence, the IBN should infer the current intent from historical ones.  

As illustrated in Fig.~\ref{network}(c), future demands correlate with historical intent through multiple dependencies:  
(i) the frequency of preceding device operations significantly affects the likelihood of subsequent actions (e.g., turning the TV on versus off);  
(ii) temporal patterns matter—the probability of switching off a light at dawn far exceeds that at dusk.  
The IBN therefore jointly captures intent–intent and time–intent correlations.  

We formalize the historical intent accessible to the IBN for user $U_j$ as $R_{U_j}^{H}(t)$:
\begin{align}
    R_{U_j}^{H}(t)=\bigl[R_{U_j}(t-l_H);\dots;R_{U_j}(t-1)\bigr],
\end{align}
where $l_H$ is the history length and each action sequence, i.e.,  
\begin{align}
    R_{U_j}(t-1)=[a_1;a_2;\dots;a_k;\dots],
\end{align}
with every action $a_k$ is defined as:
\begin{align}
\label{q1}
    a_k=[t_{a_k},d_{a_k},t_{a_k}^{d},c_{a_k}],
\end{align}
containing the action time $t_{a_k}$, device index $d_{a_k}$, running duration $t_{a_k}^{d}$, and control command $c_{a_k}$.

\subsection{User Mobility Model}
Within the IBN, user locations are neither static nor uniformly distributed around the BS.  
We therefore simulate user mobility on the continuum crowd model\cite{treuille2006continuum}, enabling an accurate evaluation of AAV trajectory planning.  
Three field functions constrain the motion.  
First, we define the user position as:  
\begin{align}
\mathbf{q}_{U_j}(t)=\bigl[x_{U_j}(t),\;y_{U_j}(t),\;0\bigr],
\end{align}
and introduce the cost field:  
\begin{align}
C(x,y,\theta)=\frac{\alpha f(x,y,\theta)+\beta+\gamma g(x,y)}{f(x,y,\theta)},
\end{align}
where $(x,y)=(x_{U_j}(t),y_{U_j}(t))$ denotes the user location, $\theta\in[-\pi,\pi]$ the moving direction, $f(x,y,\theta)$ the maximum feasible speed field jointly determined by terrain density and direction, $g(x,y)\geq 0$ the discomfort cost induced by path obstacles, and $\alpha,\beta,\gamma$ the weighting coefficients.  
The speed field $f(x,y,\theta)$ is: 
\begin{align}
f(x,y,\theta)=
\begin{cases}
f_T(x,y,\theta), & p\leq p_{\min},\\
f_T+\dfrac{p-p_{\min}}{p_{\max}-p_{\min}}, & p_{\min}<p<p_{\max},\\
f_v(x,y,\theta), & p\geq p_{\max},
\end{cases}
\end{align}
with $f_T$ the intrinsic terrain-limited speed, $f_v$ the congested-flow speed, $p$ the crowd density, and $p_{\min},p_{\max}$ experimentally calibrated constants.  

Given $C(x,y,\theta)$, each user trajectory follows the minimum-cost potential $\varphi(x,y)$ satisfying the anisotropic Eikonal equation:
\begin{align}
\|\nabla\varphi(x,y)\|=C\!\left(x,y,\frac{\nabla\varphi}{\|\nabla\varphi\|}\right).
\end{align}
Once $\varphi(x,y)$ is obtained, the instantaneous velocity becomes:
\begin{align}
[v_x,\;v_y]=-f(x,y,\theta)\frac{\nabla\varphi}{\|\nabla\varphi\|}.
\end{align}
Finally, the user position is updated as: 
\begin{align}
\label{q2}
\mathbf{q}_{U_j}(t+1)=\bigl[x_{U_j}(t)+v_x\Delta t,\;y_{U_j}(t)+v_y\Delta t,\;0\bigr]^{\mathsf{T}},
\end{align}
where $\Delta t$ is the duration of the slot.

\subsection{AAV Communication Model}

AAV inevitably encounters complex communication environments and diverse interference factors during movement. We comprehensively consider large-scale and small-scale fading to construct a communication model for the mobile process. We define the coordinates of AAV, BS, and user as $\mathbf{q}_{\mathrm{BS}} = [x_{\mathrm{BS}},\; y_{\mathrm{BS}},\; 0], \mathbf{q}_{\mathrm{V_i}}(t) = [x_{\mathrm{V_i}}(t),\; y_{\mathrm{V_i}}(t),\; z_{\mathrm{V_i}}(t)]$, and $\mathbf{q}_{\mathrm{U_j}}(t) = [x_{\mathrm{U_j}}(t),\; y_{\mathrm{U_j}}(t),\; 0]$, where $z_{\mathrm{V_i}}(t)$ is a fixed value. The 3D distance $d_{\mathrm{3D}}(t)$, horizontal distance $d_{\mathrm{2D}}(t)$, and elevation angle $\psi(t)$ between AAV and BS are calculated as follows:
\begin{align}
\label{q3}
&\ d_{\mathrm{3D}}(t) = \|\mathbf{q}_{\mathrm{V_i}}(t)-\mathbf{q}_{\mathrm{BS}}\|_2, \\
&\ d_{\mathrm{2D}}(t) = \sqrt{(x_{\mathrm{V_i}}-x_{\mathrm{BS}})^2+(y_{\mathrm{V_i}}-y_{\mathrm{BS}})^2}, \\
&\ \psi(t) = \arctan\left(\frac{z_{\mathrm{V_i}}(t)}{d_{\mathrm{2D}}(t)}\right).
\end{align}
Similarly, the 3D distance $d_{\mathrm{3D}}^u(t)$, horizontal distance $d_{\mathrm{2D}}^u(t)$, and elevation angle $\psi^u(t)$ between AAV and user are:
\begin{align}
&\ d_{\mathrm{3D}}^u(t) = \|\mathbf{q}_{\mathrm{V_i}}(t)-\mathbf{q}_{U_i}\|_2, \\
&\ d_{\mathrm{2D}}^u(t) = \sqrt{(x_{\mathrm{V_i}}-x_{U_i})^2+(y_{\mathrm{V_i}}-y_{U_i})^2}, \\
&\ \psi^u(t) = \arctan\left(\frac{z_{\mathrm{V_i}}(t)}{d_{\mathrm{2D}}^u(t)}\right).
\end{align}
The probability of LoS between AAV-user $P_{\mathrm{LoS}}^u(\psi^u)$ and AAV-BS $P_{\mathrm{LoS}}(\psi)$ is:
\begin{align}
P_{\mathrm{LoS}}(\psi) &= \frac{1}{1+a\,\exp\bigl[-b\,(\psi-a)\bigr]}, \\
P_{\mathrm{LoS}}^u(\psi^u) &= \frac{1}{1+a\,\exp\bigl[-b\,(\psi^u-a)\bigr]},
\end{align}
where $a$ and $b$ are site-measured values.

The LoS path loss for AAV-user $\mathrm{PL}_{\mathrm{LoS}}^u(d_{\mathrm{3D}}^u)$ and AAV-BS $\mathrm{PL}_{\mathrm{LoS}}(d_{\mathrm{3D}})$ is:
\begin{align}
\mathrm{PL}_{\mathrm{LoS}}(d_{\mathrm{3D}}) &= 20\log_{10}\left(\frac{4\pi d_{\mathrm{3D}}}{\lambda}\right)+\eta_{\mathrm{LoS}}\log_{10}\left(\frac{d_{\mathrm{3D}}}{d_{0}}\right), \\
\mathrm{PL}^u_{\mathrm{LoS}}(d_{\mathrm{3D}}^u) &= 20\log_{10}\left(\frac{4\pi d_{\mathrm{3D}}^u}{\lambda}\right)+\eta_{\mathrm{LoS}}\log_{10}\left(\frac{d_{\mathrm{3D}}^u}{d_{0}}\right),
\end{align}
where $d_0$ is the unit length, and $\eta_{\mathrm{LoS}}$ is an environmental measurement.

The NLoS path loss for AAV-user $\mathrm{PL}_{\mathrm{NLoS}}^u(d_{\mathrm{3D}}^u)$ and AAV-BS $\mathrm{PL}_{\mathrm{NLoS}}(d_{\mathrm{3D}})$ is:
\begin{align}
\mathrm{PL}_{\mathrm{NLoS}}(d_{\mathrm{3D}}) &= \mathrm{PL}_{\mathrm{LoS}}(d_{\mathrm{3D}})+\Delta_{\mathrm{NLoS}}, \\
\mathrm{PL}_{\mathrm{NLoS}}^u(d_{\mathrm{3D}}^u) &= \mathrm{PL}_{\mathrm{LoS}}(d_{\mathrm{3D}}^u)+\Delta_{\mathrm{NLoS}},
\end{align}
where $\Delta_{\mathrm{NLoS}}$ is a recognized empirical value.

The composite large-scale gain for AAV-user $G_{\mathrm{large}}^u(\mathrm{dB})$ and AAV-BS $G_{\mathrm{large}}(\mathrm{dB})$ is:
\begin{align}
G_{\mathrm{large}}(\mathrm{dB}) = &\ P_{\mathrm{LoS}}\bigl[-\mathrm{PL}_{\mathrm{LoS}}+X_{\mathrm{Sh,LoS}}\bigr]+ \\ \notag
&\ (1-P_{\mathrm{LoS}})\bigl[-\mathrm{PL}_{\mathrm{NLoS}}+X_{\mathrm{Sh,NLoS}}\bigr], \\
G_{\mathrm{large}}^u(\mathrm{dB})  = &\ P_{\mathrm{LoS}}^u\bigl[-\mathrm{PL}^u_{\mathrm{LoS}}+X_{\mathrm{Sh,LoS}}\bigr]+ \\ \notag
&\ (1-P^u_{\mathrm{LoS}})\bigl[-\mathrm{PL}^u_{\mathrm{NLoS}}+X_{\mathrm{Sh,NLoS}}\bigr],
\end{align}
where $X_{\mathrm{Sh,LoS}}\sim \mathcal{N}(0, \sigma_{\text{LoS}}^2)$ and $X_{\mathrm{Sh,NLoS}} \sim \mathcal{N}(0, \sigma_{\text{NLoS}}^2)$ are shadow fading in LoS and NLoS environments, respectively, both following log-normal distributions. $\sigma_{\text{LoS}}^2$ and $\sigma_{\text{NLoS}}^2$ are measured values.

For small-scale fading, the Rician model is used:
\begin{align}
h_{\mathrm{small}}(t,K)=&\ \sqrt{\frac{K}{K+1}}\,e^{j\phi_{0}}+\sqrt{\frac{1}{K+1}}\,h_{\mathrm{Ray}}(t), \\ \notag
&\ h_{\mathrm{Ray}}(t)\sim\mathcal{CN}(0,1),
\end{align}
where $\phi_0$ is the fixed LoS phase, set to 0, and $h_{\text{Ray}}(t)$ follows the complex Gaussian distribution $\mathcal{CN}(0, 1)$.

The Rician $K$ factor for AAV-user $K_{\mathrm{dB}}^u(\psi^u)$ and AAV-BS $K_{\mathrm{dB}}(\psi)$ varies with the elevation angle:
\begin{align}
K_{\mathrm{dB}}(\psi) &= \max\bigl\{K_{0}-\kappa\psi,\;K_{\mathrm{min}}\bigr\}, \\
K^u_{\mathrm{dB}}(\psi^u) &= \max\bigl\{K_{0}-\kappa\psi^u,\;K_{\mathrm{min}}\bigr\},
\end{align}
where $K_{0}, \kappa$, and $K_{\mathrm{min}}$ are channel measurement values.

The impact of small-scale fading on channel capacity is calculated as:
\begin{align}
|h_{\mathrm{small}}(t,K)|^2= &\ |\sqrt{\frac{K}{K+1}}\,e^{j\phi_{0}}+\sqrt{\frac{1}{K+1}}\,h_{\mathrm{Ray}}(t)|^2,\\ \notag
&\ h_{\mathrm{Ray}}(t)\sim\mathcal{CN}(0,1).
\end{align}

The linear SNR for AAV-user $\mathrm{SNR}_{\mathrm{lin}}^u$ and AAV-BS $\mathrm{SNR}_{\mathrm{lin}}$ is:
\begin{align}
\mathrm{SNR}_{\mathrm{lin}} = &\ 10^{\bigl(P_{\mathrm{tx}} + G_{\mathrm{T}} + G_{\mathrm{R}}-N_{0}-10\log_{10}B_{\mathrm{W}}\bigr)/10} \; / \; \\ \notag
&\ G_{\mathrm{large}} \; / \; |h_{\mathrm{small}}(t,K_{\mathrm{dB}}(\psi))|^{2}, \\
\mathrm{SNR}^u_{\mathrm{lin}} = &\ 10^{\bigl(P^u_{\mathrm{tx}} + G^u_{\mathrm{T}} + G^u_{\mathrm{R}}-N_{0}-10\log_{10}B^u_{\mathrm{W}}\bigr)/10} \; / \; \\ \notag
&\ G^u_{\mathrm{large}} \; / \; |h_{\mathrm{small}}(t,K^u_{\mathrm{dB}}(\psi))|^{2},
\end{align}
where $P_{\text{tx}}$ is the transmit power, $G_T$ and $G_R$ are antenna gains, $N_0$ is the noise power spectral density, and $B_W$ is the system bandwidth (Hz).

The channel capacity for AAV-user $C(t)^u$ and AAV-BS $C(t)$ is:
\begin{align}
C(t) &= B_{\mathrm{W}} \log_{2}\bigl(1 + \mathrm{SNR}_{\mathrm{lin}} \cdot G_{\mathrm{large}} \cdot |h_{\mathrm{small}}(t,K_{\mathrm{dB}}(\psi))|^{2}\bigr), \\
C^u(t) &= B^u_{\mathrm{W}} \log_{2}\bigl(1 + \mathrm{SNR}^u_{\mathrm{lin}} \cdot G^u_{\mathrm{large}} \cdot |h_{\mathrm{small}}(t,K^u_{\mathrm{dB}}(\psi))|^{2}\bigr).
\end{align}
The processing delay for user requirements is:
\begin{align}
\label{delay_jyz}
T_{\mathrm{delay}} &= \frac{R_{U_j}(t)}{C(t)}, \\
T^u_{\mathrm{delay}} &= \frac{R_{U_j}(t)}{C^u(t)}.
\end{align}

We have described the communication delay for AAV transferring user requirements to BS $T_{\mathrm{delay}}$ and AAV directly processing user requirements $T_{\mathrm{delay}}^u$. There is also a situation where the user's intent is not captured by AAV, and the user communicates directly with BS. This is intolerable in our designed intent-based network (IBN). The delay consumed by this direct communication is set to a large fixed value $T^{\max}_{\mathrm{delay}}$.

\section{Problem Formulation}
In this section, we unify the multiple model standards involved in the system model into the user QoE model, and further clarify the optimization goals of this paper.

\subsection{Quality of Experience Model}
We employ a generic QoE model \cite{zhou2016mobile}:
\begin{align}
\label{Q_jyz}
Q_{{U_j},{R_{{U_j}}}(t)} =&\ Q_{R_{{U_j}}(t)}(T_{{U_j},{R_{{U_j}}}(t)})- \\ \notag
&\ \int_{t_a}^{T_{{U_j},{R_{{U_j}}}}} Q_a(t_a,t)dt-Q_b(t_b,t),
\end{align}
where $ T_{{U_j},{R_{{U_j}}}(t)} $ is the actual response time of user $ U_j $'s intent $ R_{{U_j}}(t) $, representing the communication delay as shown in Eq. (\ref{delay_jyz}). $ Q_{R_{{U_j}}(t)}(T_{{U_j},{R_{{U_j}}}(t)}) $ denotes the utility function of user $ U_j $ for intent $ R_{{U_j}}(t) $, typically a time-decreasing function reflecting reduced user satisfaction with waiting time. $ Q_{R_{{U_j}}(t)}(T_{{U_j},{R_{{U_j}}}(t)}) $ is defined as \cite{li2020first}:
\begin{align}
\label{Q_1_jyz}
&\ Q_{R_{{U_j}}(t)}(T_{{U_j},{R_{{U_j}}}(t)}) = \\ \notag
&\ \begin{cases}
1-k_1 e^{-\alpha_q T_{{U_j},{R_{{U_j}}}(t)}}, & 0 \le T_{{U_j},{R_{{U_j}}}(t)} \le T_{1}\\
\beta_q-k_{2}{(T_{{U_j},{R_{{U_j}}}(t)}-T_{1})}^{\gamma_q}, & T_{1} < T_{{U_j},{R_{{U_j}}}(t)} \le T_{2}\\
\max\bigl\{0,\; \beta_q-k_{2}{(T_{2}-T_{1})}^{\gamma_q}- \\ \quad k_{3}{(T_{{U_j},{R_{{U_j}}}(t)}-T_{2})}^{2}\bigr\}, & T_{{U_j},{R_{{U_j}}}(t)} > T_{2}
\end{cases},
\end{align}
where $ T_1 $ is the first inflection point of QoE, before which QoE declines slowly. $ T_2 $, set as $ T^{\max}_{\mathrm{delay}} $, marks the second inflection point after which QoE bottoms out.
$ k_1 $ and $ k_2 $ are preset hyperparameters. $ k_1 $ determines the initial QoE decline depth, while $ k_2 $ controls the mid-stage QoE drop-off rate. $ \beta_q $ indicates the elevation height of the second segment, calculated by $ \beta_q = 1 -k_1 e^{-\alpha_q T_1} $. $\alpha_q$ and $\gamma_q$ is weighting coefficients.

The integral $\int_{t_a}^{T_{U_j,R_{U_j}}}\!Q_{a}\bigl(t_a,t\bigr)\,dt$ represents the satisfaction consumed by the user when obtaining their needs with AAV assistance within a specific time frame. $t_a$ signifies the user's upper limit for tolerating transmission delay; when $T_{U_j,R_{U_j}} \leq t_a$, the user's satisfaction remains unaffected by AAV assistance. The specific formula for calculating $Q_a$ is designed using the Beta distribution $g(x)$ \cite{mcdonald1995generalization}:
\begin{align}
&\ g(x) = \frac{x^{\alpha_b-1}(1-x)^{\beta_b-1}}{B(\alpha_b,\beta_b)}, \\
&\ B(\alpha_b,\beta_b) = \frac{\Gamma(\alpha_b)\Gamma(\beta_b)}{\Gamma(\alpha_b+\beta_b)}, \\
&\ \Gamma(n) = (n-1)!,
\end{align}
then,
\begin{align}
\label{Q_a_jyz}
Q_a(t_a,t) =
\begin{cases}
\displaystyle \frac{1}{T_2-t_a}\,g\left(\frac{t-t_a}{T_2-t_a}\right), & t_a\le t\le T_2 \\
0, & \text{otherwise}
\end{cases},
\end{align}
where $g(x)$ is the probability density function of the Beta distribution $B(\alpha_b,\beta_b)$, and $\alpha_b,\beta_b$ are hyperparameters that control the growth of the function.

The term $ Q_b(t_b,t) $ represents the QoE of a user who, after time $ t_b $, independently meets their needs via BS instead of through AAV assistance. We have designed $ Q_b(t_b,t) $ as a Sigmoid-type curve:
\begin{align}
\label{Q_b_jyz}
Q_b(t_b,t) =
\begin{cases}
0, & t \le t_b \\
\displaystyle \frac{\Bigl(\dfrac{t-t_b}{T_2-t_b}\Bigr)^n}{\Bigl(\dfrac{t-t_b}{T_2-t_b}\Bigr)^n + \Bigl(1-\dfrac{t-t_b}{T_2-t_b}\Bigr)^n}, & t_b < t < T_2 \\
1, & t \ge T_2
\end{cases},
\end{align}
where $ n $ determines the steepness of the function.

\subsection{Objective Function}
The optimization equation in this paper is transformed from Eq. \ref{Q_jyz}. Combining Eqs. \ref{Q_1_jyz}, \ref{Q_a_jyz}, and \ref{Q_b_jyz}, we determine that $ Q_{U_j,R_{U_j}(t)} \in [-2,1] $, leading to the following transformation:
\begin{align}
Q_{U_j,R_{U_j}(t)}^1 = &\ \frac{1}{3}\Bigl(Q_{R_{U_j}(t)}\bigl(T_{U_j,R_{U_j}(t)}\bigr)- \\ \notag
&\ \int_{t_a}^{T_{U_j,R_{U_j}}}\!Q_{a}\bigl(t_a,t\bigr)\,dt-Q_{b}(t_b,t) + 2\Bigr) \in [0,1].
\end{align}
However, $ Q_{U_j,R_{U_j}(t)}^1 $ only considers the satisfaction of a single user for a single intent. Thus, we define the complete optimization objective as follows:
\begin{align}
\label{Q}
\mathcal Q:\quad
\max \quad &\  \frac{1}{N_U \times T} \sum_{t \in T} \sum_{j \in N_U}  \frac{1}{3}\Bigl(Q_{R_{U_j}(t)}\bigl(T_{U_j,R_{U_j}(t)}\bigr)- \\ \notag
&\ \int_{t_a}^{T_{U_j,R_{U_j}}}\!Q_{a}\bigl(t_a,t\bigr)\,dt-Q_{b}(t_b,t) + 2\Bigr).
\end{align}

\begin{align}
% \mbox{\textit{s.t.}}\quad 
% & 0 \leq \alpha_1 \leq 1, &\tag{\ref{Q}a} \\
% & 0 \leq \alpha_2 \leq 1, &\tag{\ref{Q}b} \\
% & B > 0, &\tag{\ref{Q}c} \\
% & l_{v_i} \in \mathbb{O}, &\tag{\ref{Q}d} \\
% & I_t \geq t_{u_k,v_i}^o + t_{v_i,a}^o + t_{v_i,a}^c + t_{v_i}, \forall k, &\tag{\ref{Q}e} \\
% & I_t \geq t_{u_k,v_i}^o + t_{v_i}^c + t_{v_i}, \forall k, &\tag{\ref{Q}f} \\
% & I_t \geq t_{u_k,a}^o + t_{a}^c + t_{v_i}, \forall k, &\tag{\ref{Q}g} \\
\mbox{\textit{s.t.}}\quad
& (\ref{q1}),(\ref{q2}),(\ref{q3}-\ref{delay_jyz}),(\ref{Q_jyz}-\ref{Q_b_jyz}) &\tag{\ref{Q}a}.
\end{align}

% \subsection{Problem Statement}

\section{An Intent-Driven Framework for Autonomous Network Optimization}
This paper aims to enhance user experience in IoT by exploring the prediction accuracy of user intents and decision-making timeliness in AAV-assisted IBNs. To this end, we have developed a comprehensive prediction scheme based on the transformer architecture and proposed a joint strategy generation algorithm that leverages  MAPPO for decision-making based on prediction outcomes.

Attention mechanisms, including transformers and large-language models, have shown significant application value in the field of prediction. However, their architectures need to be adjusted to effectively predict user intents within IBNs. In IBNs, intents often consist of complex action sequences and operational objects, with individual-action intents being fickle. While prediction methods such as large models can effectively capture context, their substantial consumption of computing and storage resources is a major concern. Other attention-mechanism methods, like transformers, also face challenges related to model complexity and limited efficiency.

Firstly, we propose the HDT. By replacing complex matrix operations with HDC, HDT offers a more efficient approach. Specifically, the linear embedding layer is substituted with a Hyperdimensional encoding layer. In terms of the attention layer, Hamming distance is employed to calculate vector similarity, and approximate value search is performed using binding and unbinding operations.
Finally, considering the collaborative operation of multiple AAVs, we introduce DA-MAPPO to jointly generate individual-AAV action sequences. Unlike approaches that use a Cartesian product to handle discrete data and generate continuous data through joint strategies, our method employs separate networks to generate different actions. This approach avoids complex action combinations.

\begin{figure*}[t]
\centering 
\includegraphics[width=0.98\textwidth]{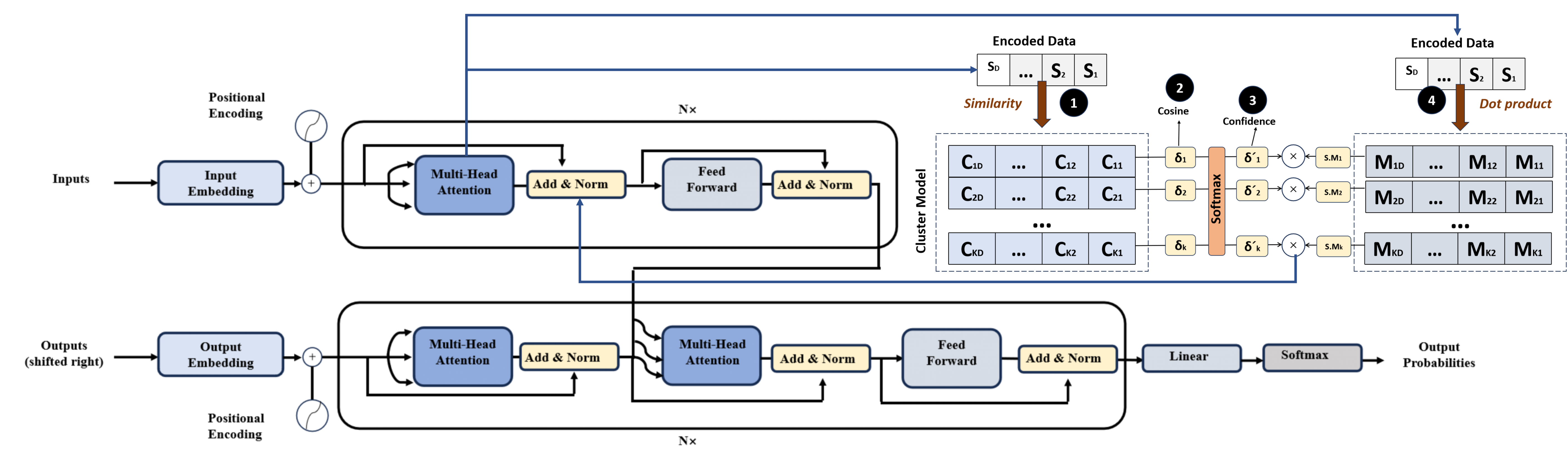}
\caption{The framework of Hyperdimensional Transformer.}
\label{HDT}
\end{figure*}

\subsection{Hyperdimensional Transformer}
In Fig.~\ref{HDT}, we aim to infer future user demands based on historical user intents. The historical intent $ R_{U_j}^H $ is used as the input. Given that $ a_k = [t_{a_k}, d_{a_k}, t_{a_k}^d, c_{a_k}] \in \mathbb{R}^{1 \times 4} $, it follows that $ R_{U_j}(t-1) = [a_1; a_2; \cdots; a_k] \in \mathbb{R}^{k \times 4} $. Consequently, the input dimension is defined as $ R_{U_j}^H(t) = [R_{U_j}(t-l_H); \cdots; R_{U_j}(t-1)] \in \mathbb{R}^{l_H \times k \times 4} $.

\textit{Hyperdimensional Encoding:} We have replaced the initial encoding layer with a Hyperdimensional encoding layer. Given $ R_{U_j}^H(t) \in \mathbb{R}^{l_H \times k \times 4} $, it can be transformed into $ R_{U_j}^H(t) = [\cdots; r_{i_r}^{\text{flatten}}; \cdots] \in \mathbb{R}^{l_H \times (k \cdot 4)} $, where $ r_{i_r} \in \mathbb{R}^{k \times 4} $ and $ r_{i_r}^{\text{flatten}} \in \mathbb{R}^{(k \cdot 4)} $. The term "flatten" refers to the process of compressing a 3D vector along a specific dimension, altering only the shape while retaining the content information. This definition applies to all subsequent instances of the "flatten" operation.

By applying the formula cited, we can obtain the preliminary Hyperdimension $ h_i $:
\begin{align}
h_{i_r} = \cos(r_{i_r}-B_i + b_{i_r}) \cdot \sin(r_{i_r}-B_{i_r}),
\end{align}
where $ B_i \sim \mathcal{N}(0, 1) $ represents randomly initialized basis vectors that ensure approximate orthogonality. This is based on Kanerva's theory:
\begin{align}
\delta(\mathbf{A}, \mathbf{B}) = \frac{\mathbf{A} \cdot \mathbf{B}}{D} \approx 0, \quad D \in [10^3, 10^6],
\end{align}
where $ \mathbf{A} $ and $ \mathbf{B} $ are randomly generated basis vectors. $ b_i \sim \mathcal{U}(0, 2\pi) $ denotes phase perturbation. After performing the vector reshape operation to eliminate the effects of "flatten", we obtain $ \mathbf{H} = (h_1, h_2, \dots, h_{l_H}) \in \mathbb{R}^{l_H \times k \times D} $, where $ D > 4 $ (for example, $ D = 10{,}000 $). The goal of this transformation is to ensure that similar inputs in the high-dimensional space yield similar vectors, preserving the structural similarity of inputs.

\textit{Positional Encoding:} To capture the temporal order of the inputs, the positional encoding matrix $ E_p \in \mathbb{R}^{l_H \times D} $ is computed as follows:
\begin{align}
E_p(t, 2i_e) = \sin\left(\frac{t}{10000^{2i_e / D}}\right), \\
E_p(t, 2i_e + 1) = \cos\left(\frac{t}{10000^{2i_e / D}}\right),
\end{align}
where $i_e$ is the information of position,
and after applying the repeat operation to replicate the $ D $ dimension $ k $ times, $ E_p \in \mathbb{R}^{l_H \times k \times D} $ aligns with the output $ \mathbf{H} $. The output of the positional encoding layer is $ Z = E_p + \mathbf{H} \in \mathbb{R}^{l_H \times k \times D} $, which integrates both positional and content information and the $i$-th encoding layer is $Z_i$.

\textit{Layer $ i $ Attention Mechanism:} Taking the $ i $-th layer as an example, the input to this layer is the output from the $ (i-1) $-th layer, denoted as $ Z_{i-1} \in \mathbb{R}^{l_H \times k \times D} $. Within the same layer, the $ Q $, $ K $, and $ V $ matrices share the same input $ Z_{i-1} $. Due to the hyperdimensional nature of $ Z_{i-1} $, the dimensions of the learned matrices for $ Q $, $ K $, and $ V $ are adjusted to $ W_Q, W_K, W_V \in \mathbb{R}^{D \times D} $, resulting in:
\begin{align}
&\ Q = Z_{i-1}^{\text{flatten}} \cdot W_Q, \quad K = Z_{i-1}^{\text{flatten}} \cdot W_K, \\ \notag
&\ V = Z_{i-1}^{\text{flatten}} \cdot W_V \in \mathbb{R}^{(l_H \cdot k) \times D}, \quad Z_{i-1}^{\text{flatten}} \in \mathbb{R}^{(l_H \cdot k) \times D}.
\end{align}
where $Z_{i-1}^{\text{flatten}} \in \mathbb{R}^{(l_H\cdot k)\times D}$.

The similarity calculation in the Hyperdimensional space differs from the original attention mechanism. The attention weights, or similarity $ \delta(Q, K) $, are computed using the Hamming distance. The aggregated output is expressed as:
\begin{align}
Z_i^{\text{output}} = \sum \text{softmax}(\delta(Q, K)) \cdot V,
\end{align}
and after reshaping, the output of the $ i $-th layer is $ Z_i^{\text{output}} \in \mathbb{R}^{l_H \times k \times D} $.

\textit{Projection Matrix:} To produce the final prediction with the desired dimensionality, a final linear layer with $ W_{\text{out}} \in \mathbb{R}^{D \times 4} $ and $ b_{\text{out}} \in \mathbb{N}^4 $ is applied:
\begin{align}
Y_{\text{all}}' = Z_{\text{output}}^{\text{flatten}} \cdot W_{\text{out}} + b_{\text{out}} \in \mathbb{R}^{(l_H \cdot k) \times 4}, \quad Z_{\text{output}}^{\text{flatten}} \in \mathbb{R}^{(l_H \cdot k) \times D},
\end{align}
where $ Z_{\text{output}} \in \mathbb{R}^{l_H \times k \times D} $ is the vector after passing through all attention layers, residual connections, LayerNorm, position-wise feed-forward networks (FFN), and a second residual connection with LayerNorm. After reshaping, $ Y_{\text{all}}' \in \mathbb{R}^{(l_H \cdot k) \times 4} $ becomes $ Y_{\text{all}} \in \mathbb{R}^{l_H \times k \times 4} $, which is the network's final output. We extract the required prediction length $ N_{\text{output}} $:
\begin{align}
\hat{Y} = Y_{\text{all}}[-N_{\text{output}}:, :, :].
\end{align}
The loss is computed using standard Mean Squared Error (MSE):
\begin{align}
Loss = MSE(\overline{Y},\hat{Y}).
\end{align}
where $\overline{Y}$ is the true value.

\begin{figure*}[t]
\centering 
\includegraphics[width=1\textwidth]{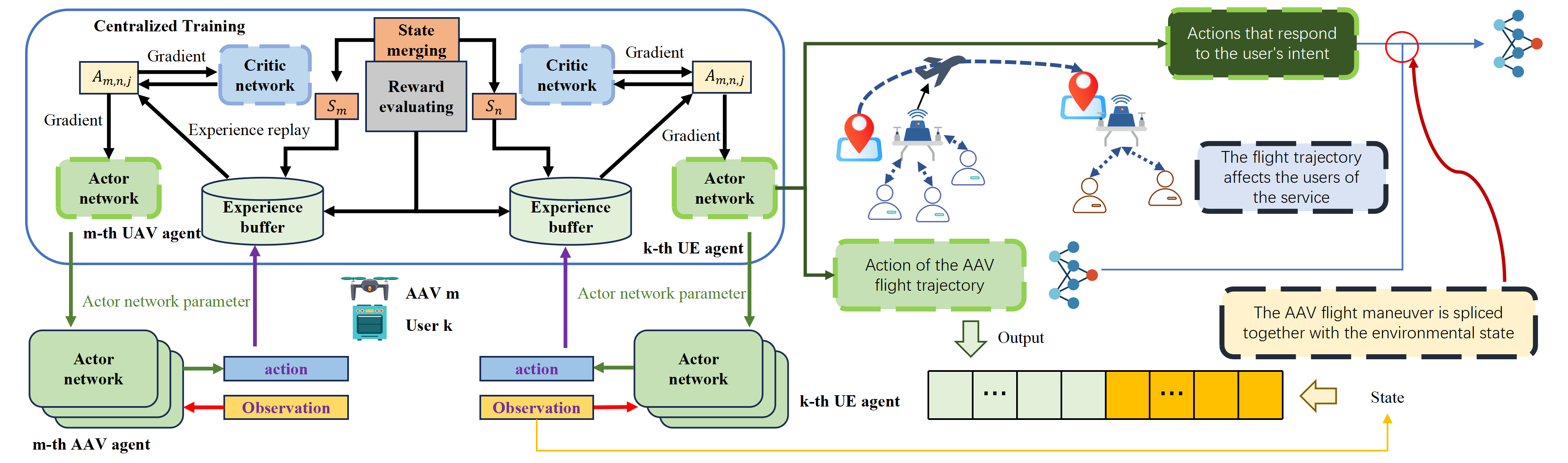}
\caption{The framework of Double Actions based Multi Agent Proximal Policy Optimization.}
\label{JA}
\end{figure*}

\subsection{Double Actions based Multi Agent Proximal Policy Optimization}
In this section, after predicting future user intents using HDT, we propose the use of DA-MAPPO to coordinate multiple AAVs in an IBN.

Fig.~\ref{JA} illustrates the DA-MAPPO architecture, which comprises three modules: environment exploration, policy optimization, and double actions. Specifically, the environment exploration follows the MAPPO procedure, integrating each agent's development strategy. The policy optimization module utilizes the clipped objective function of PPO, namely the problem AAA in this paper, to ensure policy updates within a constrained range. Moreover, the policy optimization module concatenates the prediction results of HDT to the beginning of its input vector. Consequently, the policy optimization is guided by both QoE factors and subjective intents, endowing the generated strategy with inherent intent perception and the ability to generate policies based on historical user intents.The double action module simultaneously considers AAV trajectory planning and intent response, treating trajectory planning actions as prior conditions for intent response and sequentially outputting both. This approach achieves simultaneous AAV decision-making while considering the correlation between actions.

\textit{MDP design:}
We model the joint optimization problem defined in Eq. \ref{Q} as a Markov Decision Process (MDP). The key components of this MDP are defined as follows:

\begin{itemize}
    \item \textbf{State Space:} According to the network model in the system model, the state space of each AAV in the IBN includes its position, the users it serves, each user's historical intents, communication status with users, and communication status with the BS. The complete global state space is represented as:
    \begin{align}
        &\ S_t = [\mathbf{q}_{\mathrm{BS}}, s_t^{V_1}, \cdots, s_t^{V_{N_v}}], \\
        &\ s_t^{V_i} = [\mathbf{q}_{\mathrm{V_i}}, \overbrace{J_t^{V_i,U_{j_1}}, \cdots, J_t^{V_i,U_{j_2}}}^{N_l}],\\
        &\ j_1,j_2 \in \{1,2,\cdots,N_u\}, \nonumber\\
        &\ J_t^{V_i,U_j} = [R_{U_j}^H, \mathbf{q}_{\mathrm{U_j}}],
    \end{align}
    where $ J_t^{V_i,U_j} $ denotes the information of user $ U_j $ associated with AAV $ V_i $.

    \item \textbf{Action Space:} Since each AAV has the same form of actions, we present the action space for AAV $ V_i $. Considering the user movement patterns in the environment, the action sequence of the AAV must first include its flight direction and speed. Additionally, the AAV must select or discard the information predicted by HDT to further refine future user intents. If the stored intents do not match the user's intent, the AAV must also forward the user's request to the BS. Therefore, the action space for AAV $ V_i $ is defined as:
    \begin{align}
        A^{V_i}_t = [v_x^f, v_y^f, m^{V_i}_t, m^{V_i,BS}_t],
    \end{align}
    where $v_x^f, v_y^f$ are the speed of AAV $V_i$ in $x$ and $y$, $ m^{V_i}_t \in \mathbb{R}^{N_{output}} $ represents the selection of intents from the prediction results $ \hat{Y} $, with $ m^{V_i}_t \in \{0,1\} $ (1 for retaining the intent, 0 for discarding it). $ m^{V_i,BS}_t \in \mathbb{R}^{N_l} $ represents the action of forwarding user requests to the BS, with $ m^{V_i,BS}_t \in \{0,1\} $ (1 for forwarding to the BS, 0 for stopping responding to the user's intent).

    \item \textbf{Reward:} In the MAPPO framework, after each action, the reward obtained by each AAV must be stored. Based on Eq. \ref{Q}, the reward obtained by AAV $ V_i $ at time $ t $ is defined as $ G_t^{V_i} $:
    \begin{align}
        G_t^{V_i} = &\ \frac{1}{N_U} \sum_{j \in \{j_1, \cdots, j_2\}} \frac{1}{3} \Bigl( Q_{R_{U_j}(t)}\bigl(T_{U_j,R_{U_j}(t)}\bigr)- \\ \notag
        &\ \int_{t_a}^{T_{U_j,R_{U_j}}} Q_a(t_a,t) dt-Q_b(t_b,t) + 2 \Bigr).
    \end{align}
    The global reward is $ G_t^{all} = [G_t^{V_1}, \cdots, G_t^{V_{N_v}}] $.
\end{itemize}

\textit{Double Actions:} In DA-MAPPO, double actions are reflected in the composition of the policy network. We first initialize the policy network $ \pi_{\theta_i} $ and value network $ \pi_{\phi_i} $ for each agent $ V_i $. The action $ A^{V_i}_t $ is divided into two parts: $ A^{1,V_i}_t = [v_x^f, v_y^f] $ for AAV flight trajectory and $ A^{2,V_i}_t = [m^{V_i}_t, m^{V_i,BS}_t] $ for user intent response. Correspondingly, the policy network consists of two parts: $ \pi_{\theta_i}^1 $ and $ \pi_{\theta_i}^2 $. $ \pi_{\theta_i}^1 $ takes the observation vector $ s_t^{V_i} $ as input, processes it through multiple fully connected layers, and outputs the probability distribution $ \mathcal{F}_{\pi_{\theta_i}^1}(A_t^{1,V_i}|s_t^{V_i}) $ (with $ v_x^f, v_y^f $ in a continuous action space). $ \pi_{\theta_i}^2 $ takes the concatenation of the observation vector $ s_t^{V_i} $ and the action probability distribution output by $ \pi_{\theta_i}^1 $ as input, resulting in the output $ \mathcal{F}_{\pi_{\theta_i}^2}(A_t^{2,V_i}|A_t^{1,V_i}, s_t^{V_i}) $.

\textit{Environment Exploration:} The AAV begins interacting with the environment, and the obtained information is stored in the replay buffer for subsequent network updates. Once a sufficient amount of trajectory data is collected, we proceed to the advantage function and target value function calculation phase. For each time step $ t $ and each agent $ i $, we compute the advantage function $ O_{t}^{V_i} $ and the target value function $ \hat{\pi_{\phi_i}} $. The advantage function $ O_{t}^{V_i} $ measures how much better or worse a specific action is compared to the average action in the current state and is calculated as:
\begin{align}
    O_{t}^{V_i} = G_t^0-\pi_{\phi_i}(s_t^{V_i}),
\end{align}
where $ G_t^0 $ is the cumulative discounted reward from time step $ t $ to $ t+n $, and $ n $ is the advantage estimation step length. The target value function $ \hat{\pi_{\phi_i}} $ is used to update the value network and is calculated as:
\begin{align}
    \hat{\pi_{\phi_i}} = G_t^0 + \gamma_p^n \pi_{\phi_i}(s_t^{V_i}),
\end{align}
where $ \gamma_p $ is the discount factor.

\textit{Policy Optimization:} In this part, we update the parameters of the policy network and value network. For each agent $ i $, we sample a batch of data from the trajectory buffer. Since the action network is divided into two independent parts, the probability distribution $ \mathcal{F}_{\pi_{\theta_i}} $ of the current policy network on the sampled data is calculated by combining the probability distributions $ \mathcal{F}_{\pi_{\theta_i}^1} $ and $ \mathcal{F}_{\pi_{\theta_i}^2} $ from the two networks:
\begin{align}
    \mathcal{F}_{\pi_{\theta_i}}(A_t^{V_i}|s_t^{V_i}) = &\ \mathcal{F}_{\pi_{\theta_i}}(A_t^{1,V_i}, A_t^{2,V_i}|s_t^{V_i}) = \mathcal{F}_{\pi_{\theta_i}^1}(A_t^{1,V_i}|s_t^{V_i}) \times  \\ \notag
    &\ \mathcal{F}_{\pi_{\theta_i}^2}(A_t^{2,V_i}|A_t^{1,V_i}, s_t^{V_i}).
\end{align}
Similarly, the probability distribution $ \mathcal{F}_{\pi_{\theta_i}^{old}} $ of the old policy network (the policy used during sampling) is calculated as:
\begin{align}
    \mathcal{F}_{\pi_{\theta_i}^{old}}(A_t^{V_i}|s_t^{V_i}) = &\ \mathcal{F}_{\pi_{\theta_i}^{old}}(A_t^{1,V_i}, A_t^{2,V_i}|s_t^{V_i}) =\mathcal{F}_{\pi_{\theta_i}^{1,old}}(A_t^{1,V_i}|s_t^{V_i}) \times  \\ \notag
    &\ \mathcal{F}_{\pi_{\theta_i}^{2,old}}(A_t^{2,V_i}|A_t^{1,V_i}, s_t^{V_i}).
\end{align}

Next, we construct the PPO objective function for the policy network:
\begin{align}
    L^{\text{CLIP}}(\pi_{\theta_i}) = &\ \mathbb{E}\left[\min(\frac{\mathcal{F}_{\pi_{\theta_i}}(A_t^{V_i}|s_t^{V_i})}{\mathcal{F}_{\pi_{\theta_i}^{old}}(A_t^{V_i}|s_t^{V_i})} O_t^{V_i}, \right.\\
    &\ \left. \text{clip}(\frac{\mathcal{F}_{\pi_{\theta_i}}(A_t^{V_i}|s_t^{V_i})}{\mathcal{F}_{\pi_{\theta_i}^{old}}(A_t^{V_i}|s_t^{V_i})}, 1-\epsilon, 1+\epsilon) O_t^{V_i})\right],
\end{align}
where $ \epsilon $ is the PPO clipping range.

This objective function ensures stable policy updates by limiting the range of policy updates, with the PPO clipping range $ \epsilon $. Meanwhile, the loss function for the value network is calculated as:
\begin{align}
    L^{\text{VF}}(\pi_{\phi_i}) = \mathbb{E}\left[(\pi_{\phi_i}(s_t^{V_i})-\hat{\pi_{\phi_i}})^2\right],
\end{align}
which is used to update the value network to more accurately predict the target value function. Additionally, we consider the policy entropy:
\begin{align}
    H(\mathcal{F}_{\pi_{\theta_i}}) = -\mathbb{E}\left[\sum_a \mathcal{F}_{\pi_{\theta_i}}(A_t^{V_i}|s_t^{V_i}) \log \mathcal{F}_{\pi_{\theta_i}}(A_t^{V_i}|s_t^{V_i}) \right],
\end{align}
to encourage policy exploration. The final total loss function is:
\begin{align}
    L(\pi_{\theta_i}, \pi_{\phi_i}) = L^{\text{CLIP}}(\pi_{\theta_i}) + c_1 H(\mathcal{F}_{\pi_{\theta_i}}) + c_2 L^{\text{VF}}(\pi_{\phi_i}),
\end{align}
where $ c_1 $ is the policy entropy coefficient and $ c_2 $ is the value function loss coefficient.

Finally, we employ stochastic gradient descent (SGD) to update the parameters $ \theta_i $ and $ \phi_i $ of the policy network and value network, respectively, to minimize the total loss function.

\begin{table}[t!]
\centering
\caption{Key Parameters of Mobility and Communication Models}
\label{para}
\begin{tabularx}{0.5\textwidth}{@{}llll@{}}
\toprule
\multicolumn{2}{c}{\textbf{Mobility Model}} & 
\multicolumn{2}{c}{\textbf{Communication Model (Sub-6 GHz)}} \\
\midrule
Parameter & Value & Parameter & Value \\
\midrule
$p_{\min}$ & $0.5$ ped/m$^2$ & $a$ (LoS/NLoS weight) & $0.6$ \\
$p_{\max}$ & $5$ ped/m$^2$ & $b$ (LoS/NLoS weight) & $0.2$ \\
$\alpha$ & $1.0$ & $d_0$ & $1$ m \\
$\beta$ & $0.5$ & $\eta_{\text{LoS}}$ & $2.0$ \\
$\gamma$ & $2.0$ & $\Delta_{\text{NLoS}}$ & $7.8\pm1.1$ dB \\
$f_T$ & $1.3$ m/s & $\sigma_{\text{Sh,LoS}}$ & $3$ dB \\
$\Delta t$ & $0.2$ s & $\sigma_{\text{Sh,NLoS}}$ & $8$ dB \\
\midrule
\multicolumn{2}{c}{\textbf{Antenna \& RF Parameters}} & \multicolumn{2}{c}{\textbf{Q-Function Parameters}} \\
\midrule
Parameter & Value & Parameter & Value \\
\midrule
$P_{\text{tx}}$ & $23$--$30$ dBm & $T_1$ & $3.2$ s \\
$G_T$ & $3$ dBi & $T_2$ & $9.5$ s \\
$G_R$ & $3$ dBi & $\alpha_q$ & $0.65$ \\
$B_w$ & $20$ MHz & $k_1$ & $0.18$ \\
$N_0$ & $-169$ dBm/Hz & $\gamma_q$ & $1.45$ \\
$\varphi_0$ & $0^\circ$ & $k_2$ & $0.023$ \\
$K_0$ & $15$ dB & $k_3$ & $0.0078$ \\
$\kappa$ & $0.4$ dB/deg & $\alpha_b$ & $4$ \\
$K_{\min}$ & $-20$ dB & $\beta_b$ & $1$ \\
-- & -- & $n$ & $5$ \\
\bottomrule
\end{tabularx}
\end{table}

\section{Experimental verification and analysis}
In this section, we explore the dataset utilized in the experiments, examine the baseline algorithms, and outline the experimental parameters. Additionally, we conduct an in-depth evaluation of the performance of DA-MAPPO with HDT across diverse experimental setups.

\subsection{Experimental Scenes and Datasets}
To underscore the effectiveness of our proposed DA-MAPPO with HDT, we conducted experimental comparisons with a selection of representative methods.
\begin{itemize}
\item \textbf{Prediction Models:}
\begin{itemize}
\item Transformer \cite{han2021transformer}:  leverages parallel computation and self-attention mechanisms to efficiently model sequences and capture long-range dependencies, making it highly effective for tasks like language modeling and intent prediction.
\item LSTM (Long Short-Term Memory) \cite{graves2012long}: utilizes memory cells and gating mechanisms to learn long-term dependencies, though with higher computational overhead, enabling it to handle sequential data with temporal patterns.
\end{itemize}
\item \textbf{Optimization Models:}
\begin{itemize}
\item DQN (Deep Q-Network)\cite{huang2020deep}: employs Q-learning and experience replay to approximate optimal policies but can be unstable due to overestimation of Q-values.
\item DDQN (Double Deep Q-Network)\cite{van2016deep}: improves upon DQN by using a target network to reduce Q-value overestimation, enhancing stability and performance in reinforcement learning tasks.
\item PPO (Proximal Policy Optimization)\cite{schulman2017proximal}: ensures stable training and high sample efficiency through a clipped surrogate objective function, making it suitable for complex decision-making tasks requiring reliable policy updates.
\end{itemize}
\end{itemize}

In the experiments, the dataset \cite{jeon2022accurate} were divided into training, validation, and test sets in a 7:2:1 ratio. The time steps for both historical and predicted traffic flow were set to 20.
In order to comprehensively demonstrate the predictive capabilities of DA-MAPPO and HDT, we conducted experiments in real mobile communication scenarios, with detailed parameters presented in Table \ref{para}.

\begin{figure}%[]
 \centering
  \includegraphics[width=0.49\textwidth]{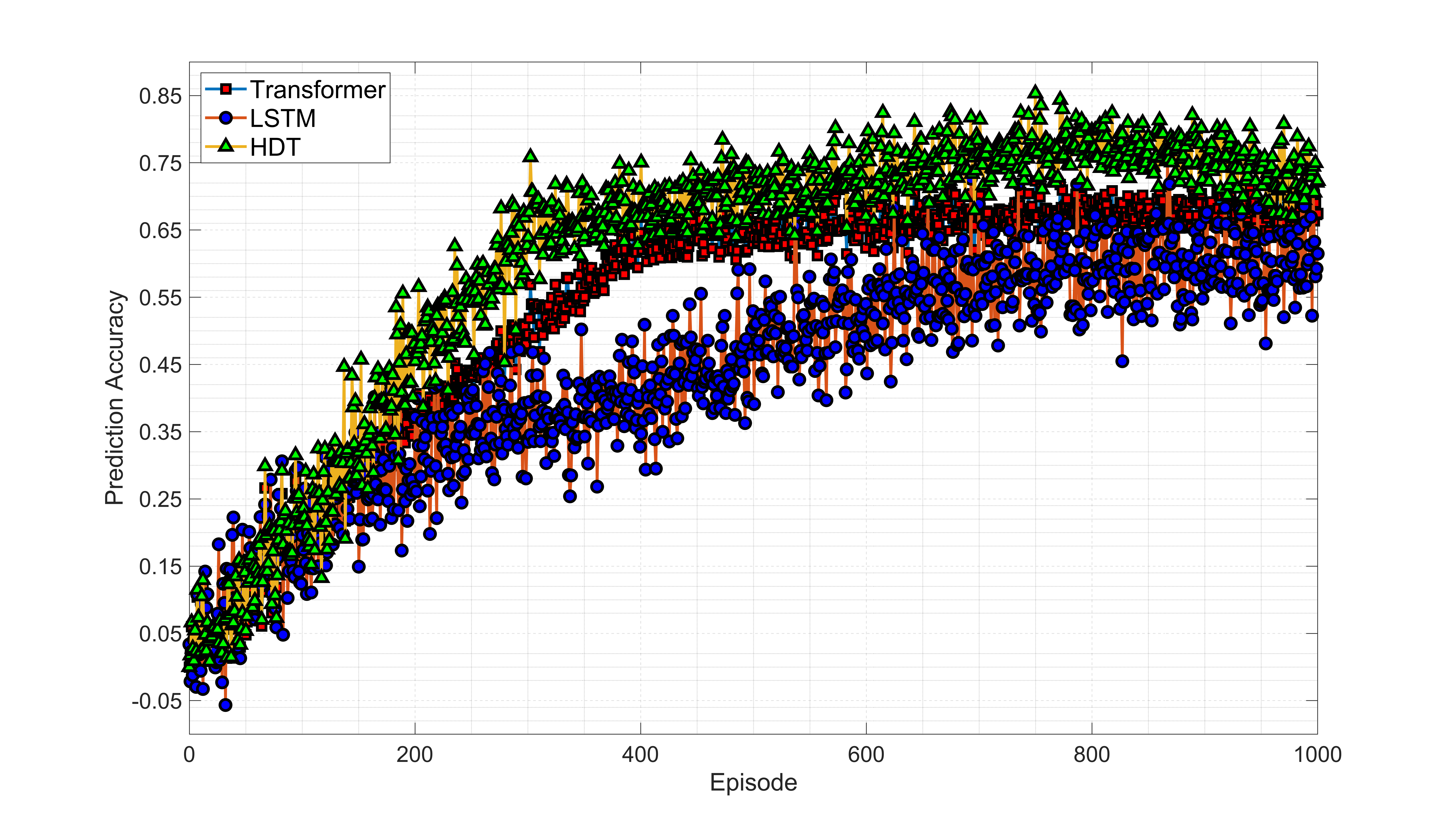}
   \caption{The training trends of intent prediction algorithms.}
   \label{p_p}
\end{figure}

\begin{figure}%[]
 \centering
  \includegraphics[width=0.49\textwidth]{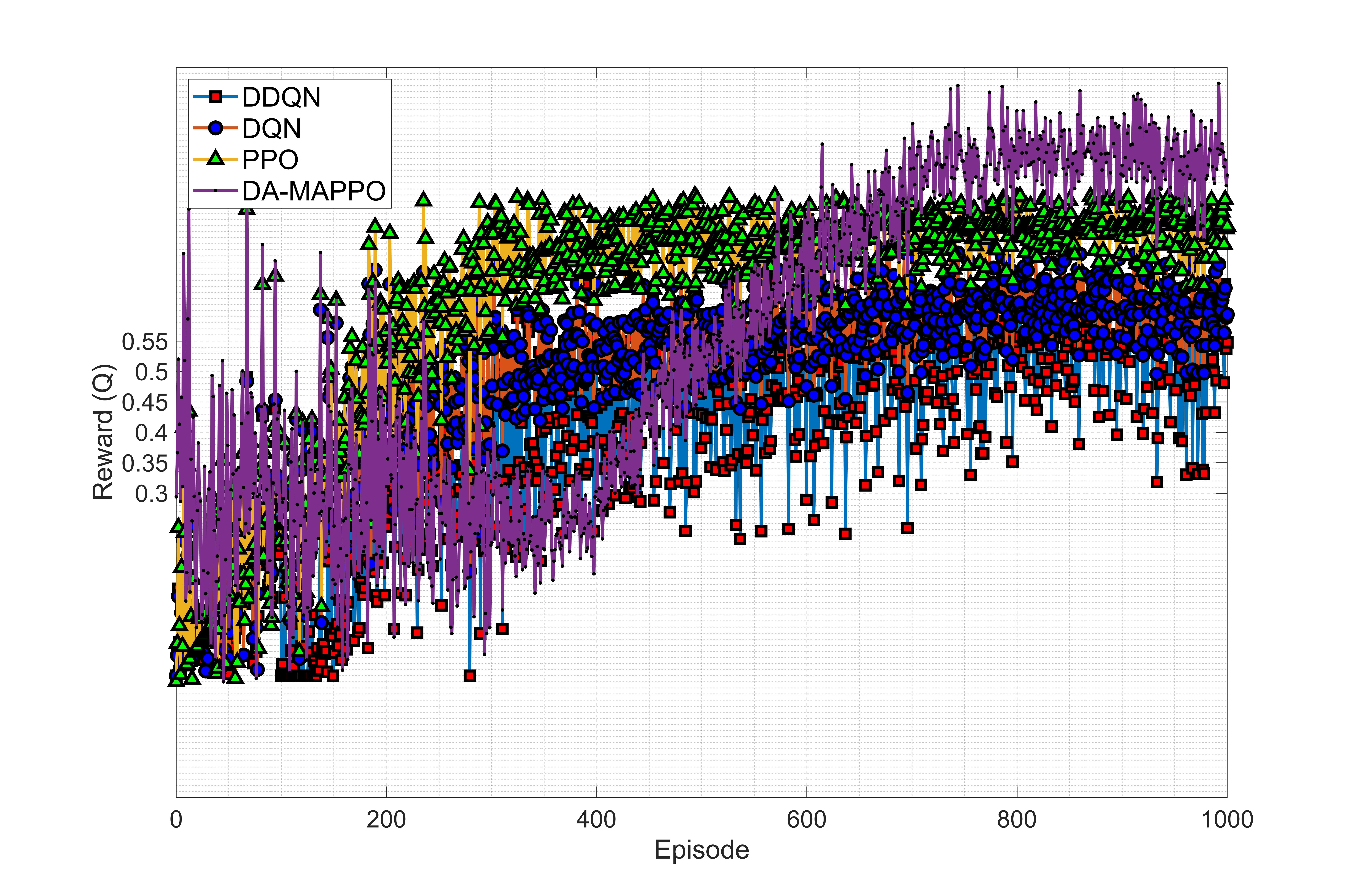}
   \caption{The training trends of network optimization algorithms.}
   \label{drl_p}
\end{figure}

\begin{figure}%[]
 \centering
  \includegraphics[width=0.49\textwidth]{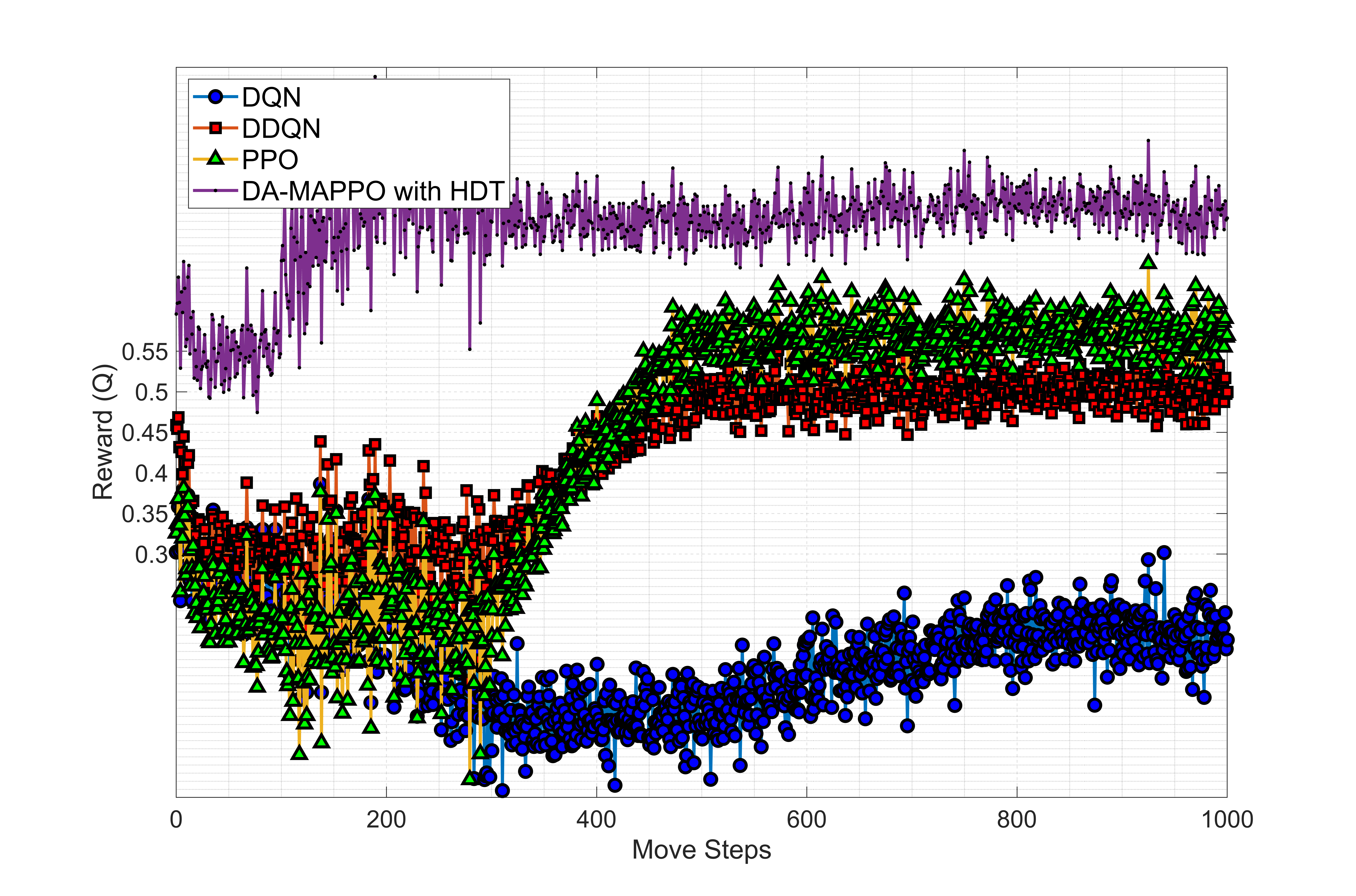}
   \caption{Performance of different decision-making frameworks in user movement scenarios.}
   \label{drl_f}
\end{figure}

\begin{figure*}[t]
	\begin{minipage}[t]{0.18\textwidth}
		\centering
		{
			\includegraphics[width=\textwidth]{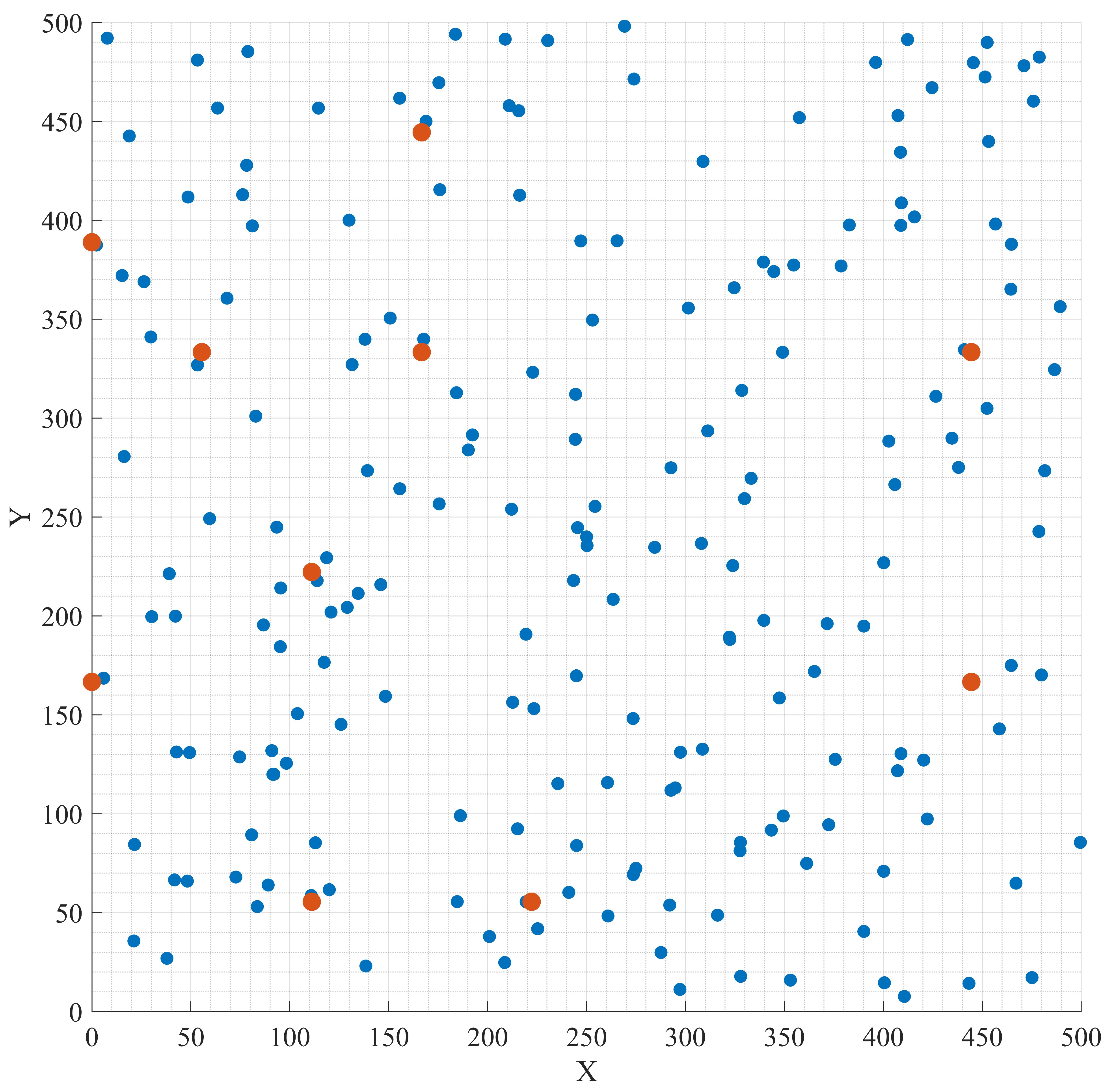}
			%\label{Fig4_a:hswish}
		}	
	\end{minipage}
	\begin{minipage}[t]{0.18\textwidth}
		\centering
		{
			\includegraphics[width=\textwidth]{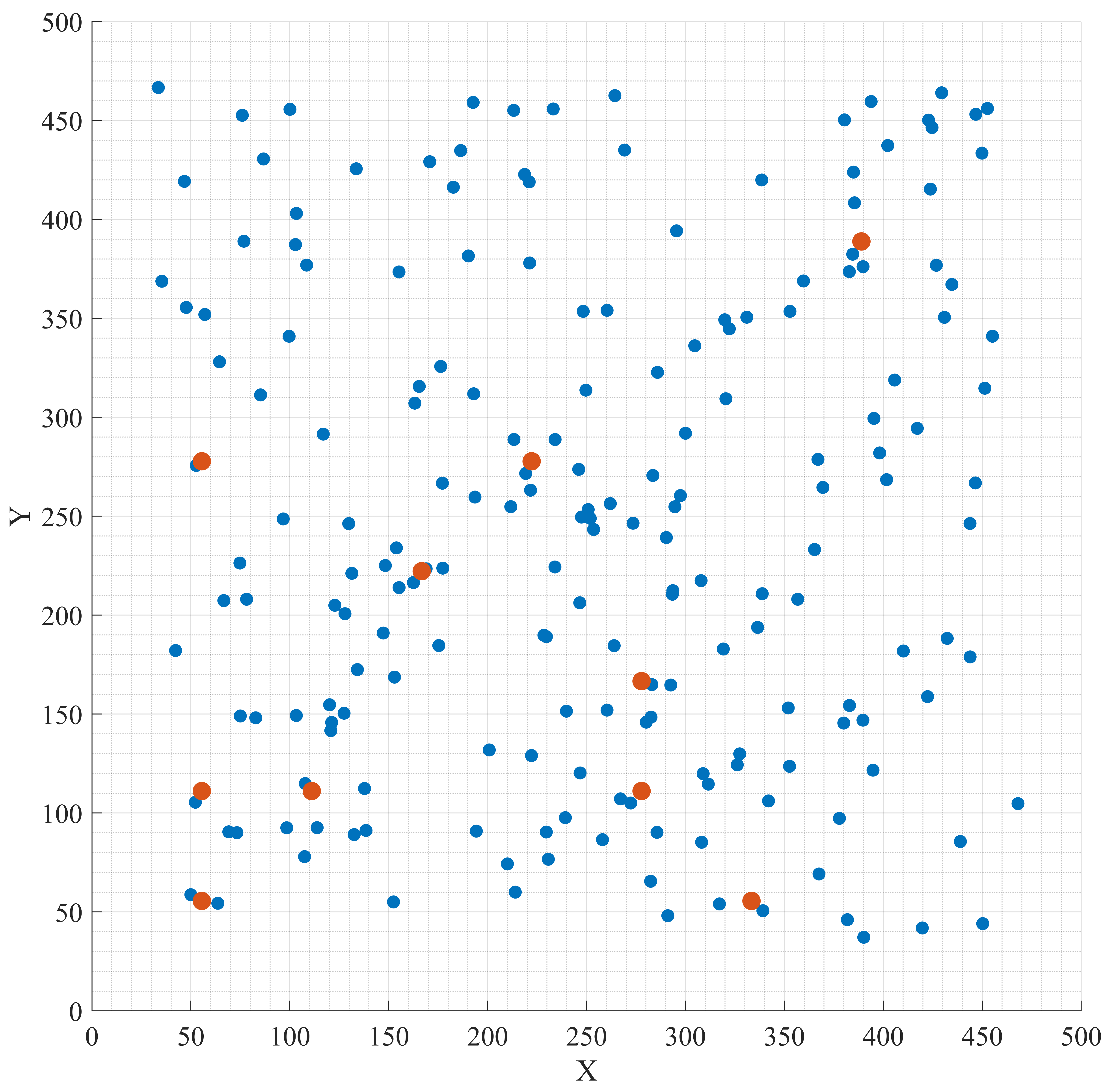}
			%\label{Fig4_a:hswish}
		}	
	\end{minipage}
    \begin{minipage}[t]{0.18\textwidth}
		\centering
		{
			\includegraphics[width=\textwidth]{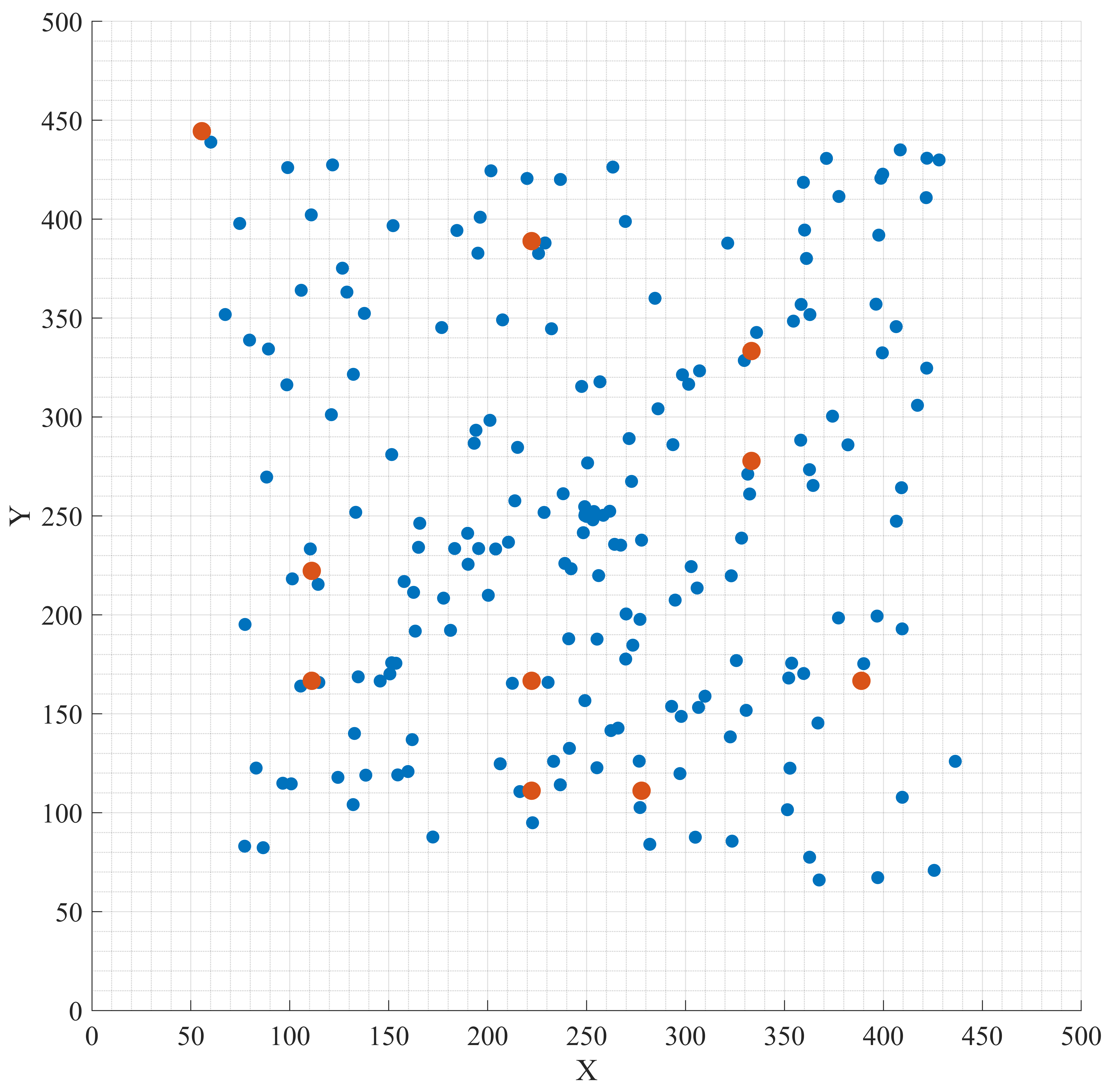}
			%\label{Fig4_a:hswish}
		}	
	\end{minipage}
    \begin{minipage}[t]{0.18\textwidth}
		\centering
		{
			\includegraphics[width=\textwidth]{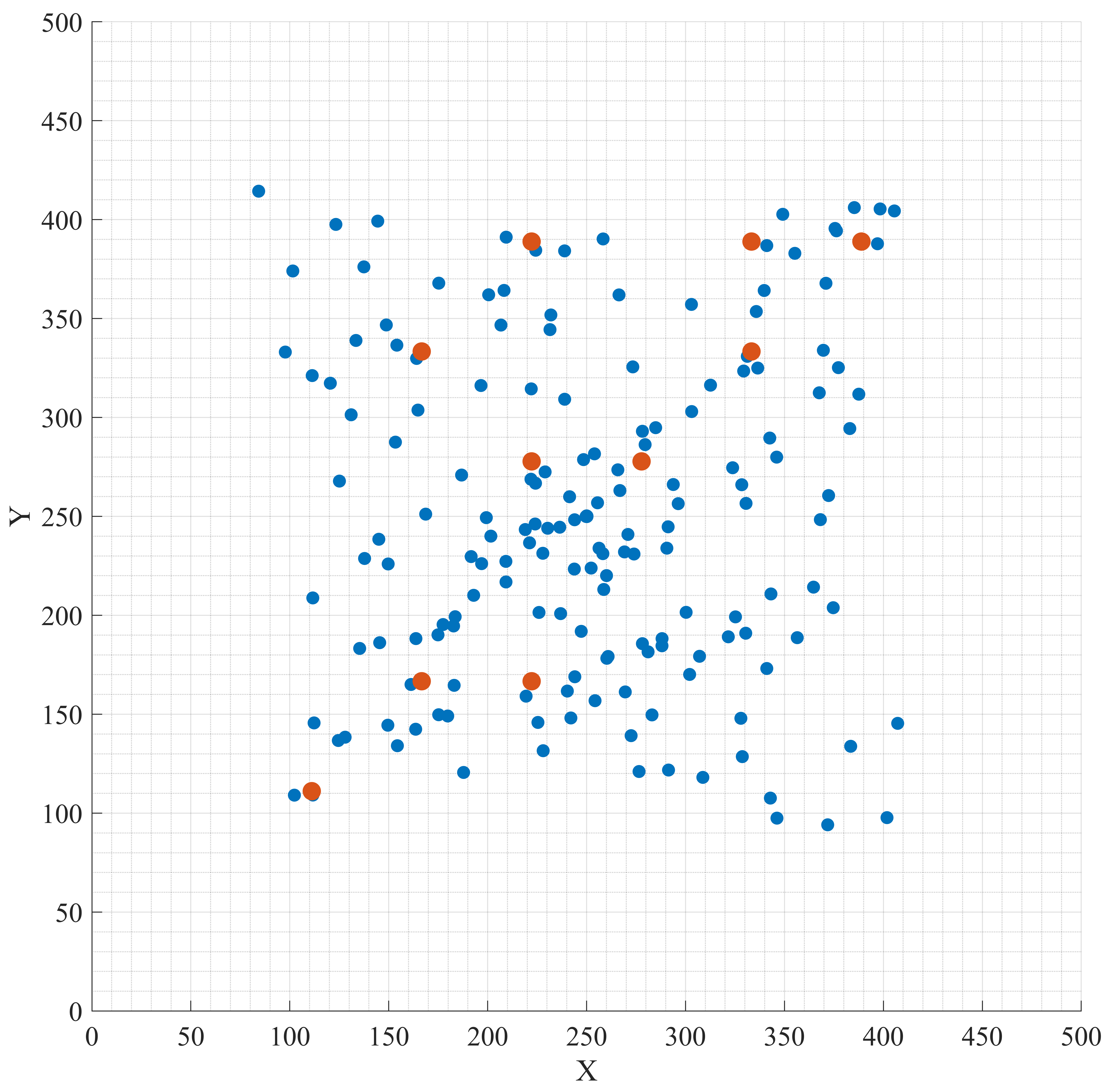}
			%\label{Fig4_a:hswish}
		}	
	\end{minipage}
    \begin{minipage}[t]{0.18\textwidth}
		\centering
		{
			\includegraphics[width=\textwidth]{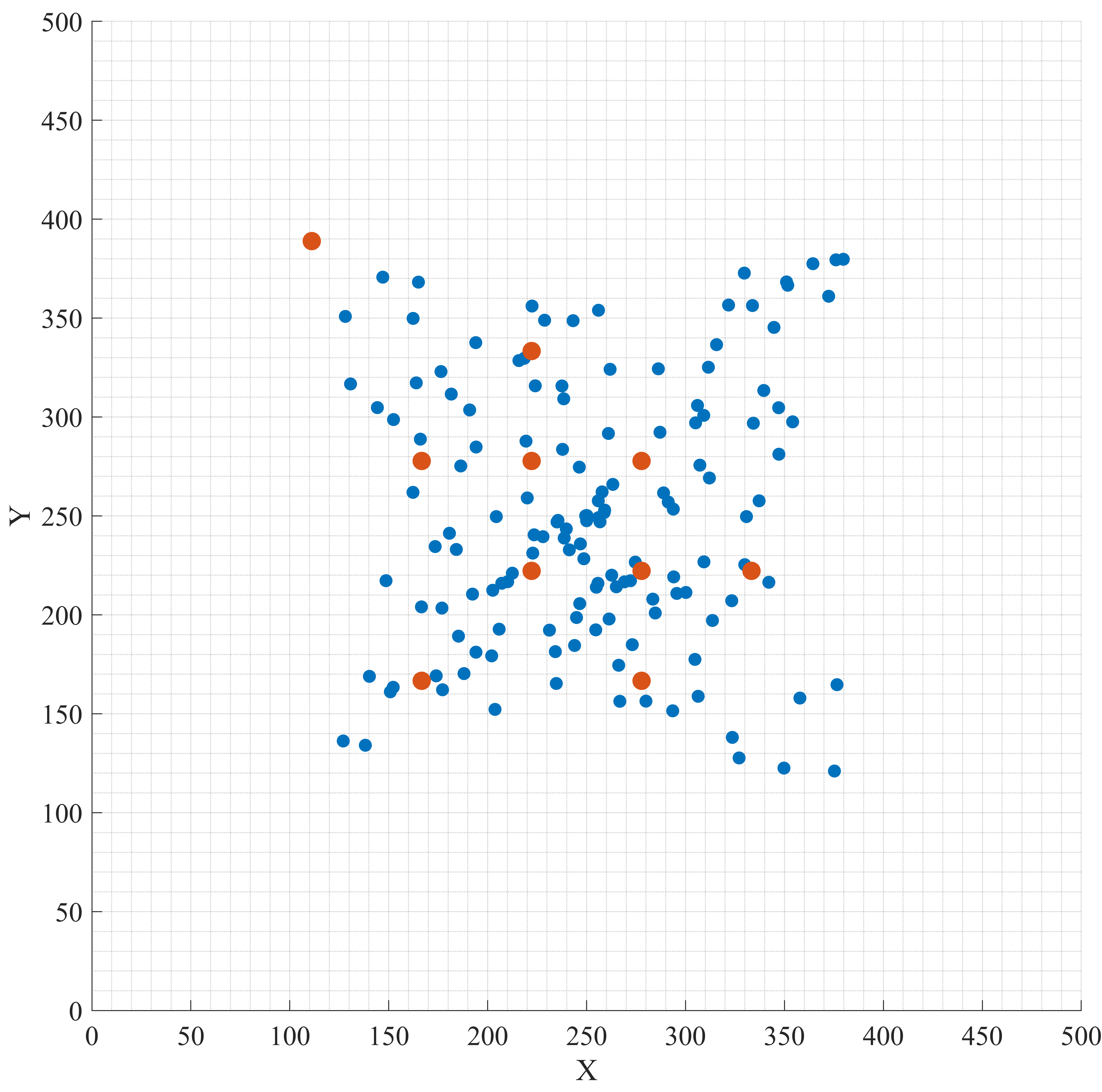}
			%\label{Fig4_a:hswish}
		}	
	\end{minipage} 

\begin{minipage}[t]{0.18\textwidth}
		\centering
		{
			\includegraphics[width=\textwidth]{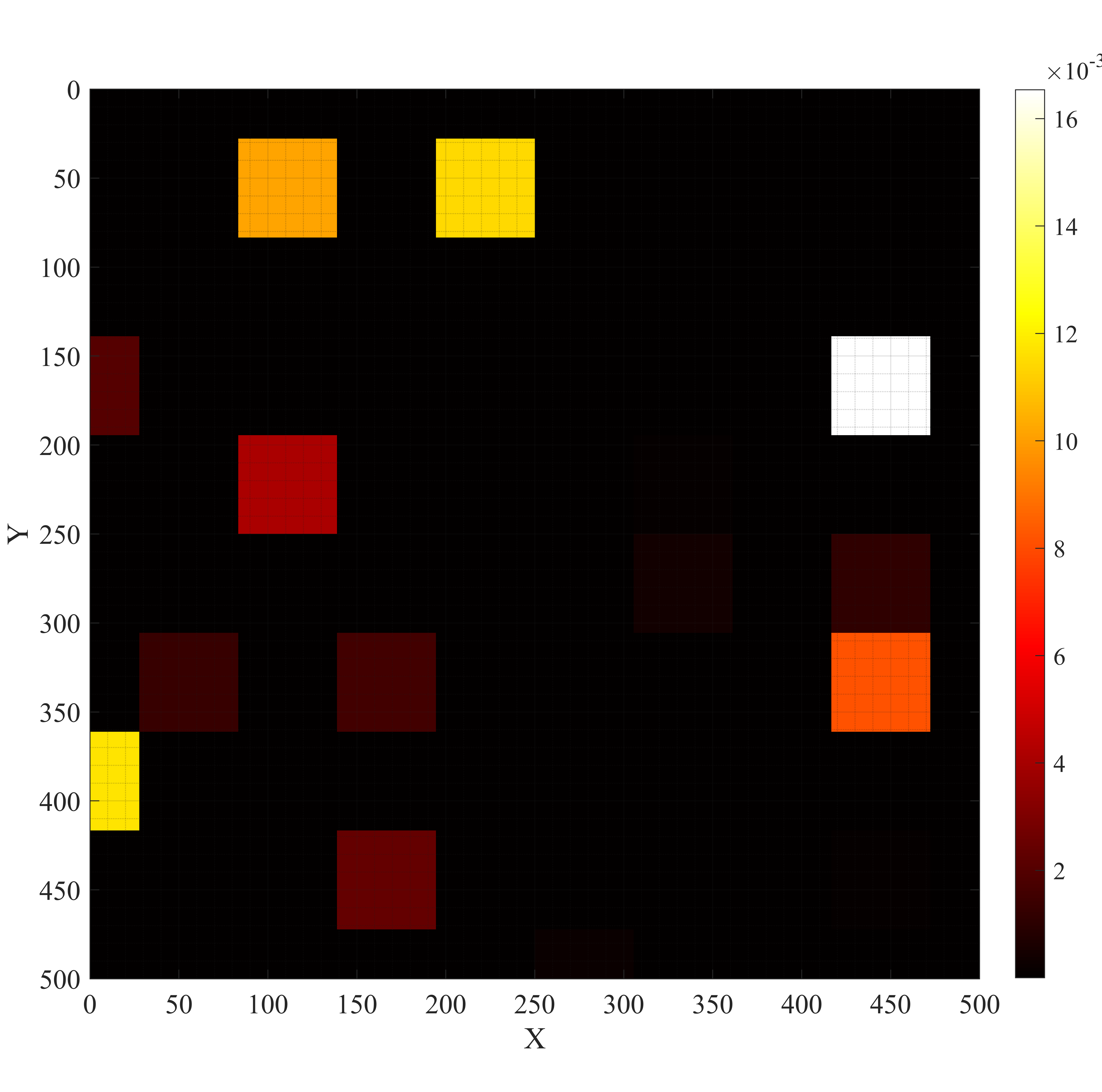}
			%\label{Fig4_a:hswish}
		}	
	\end{minipage}
	\begin{minipage}[t]{0.18\textwidth}
		\centering
		{
			\includegraphics[width=\textwidth]{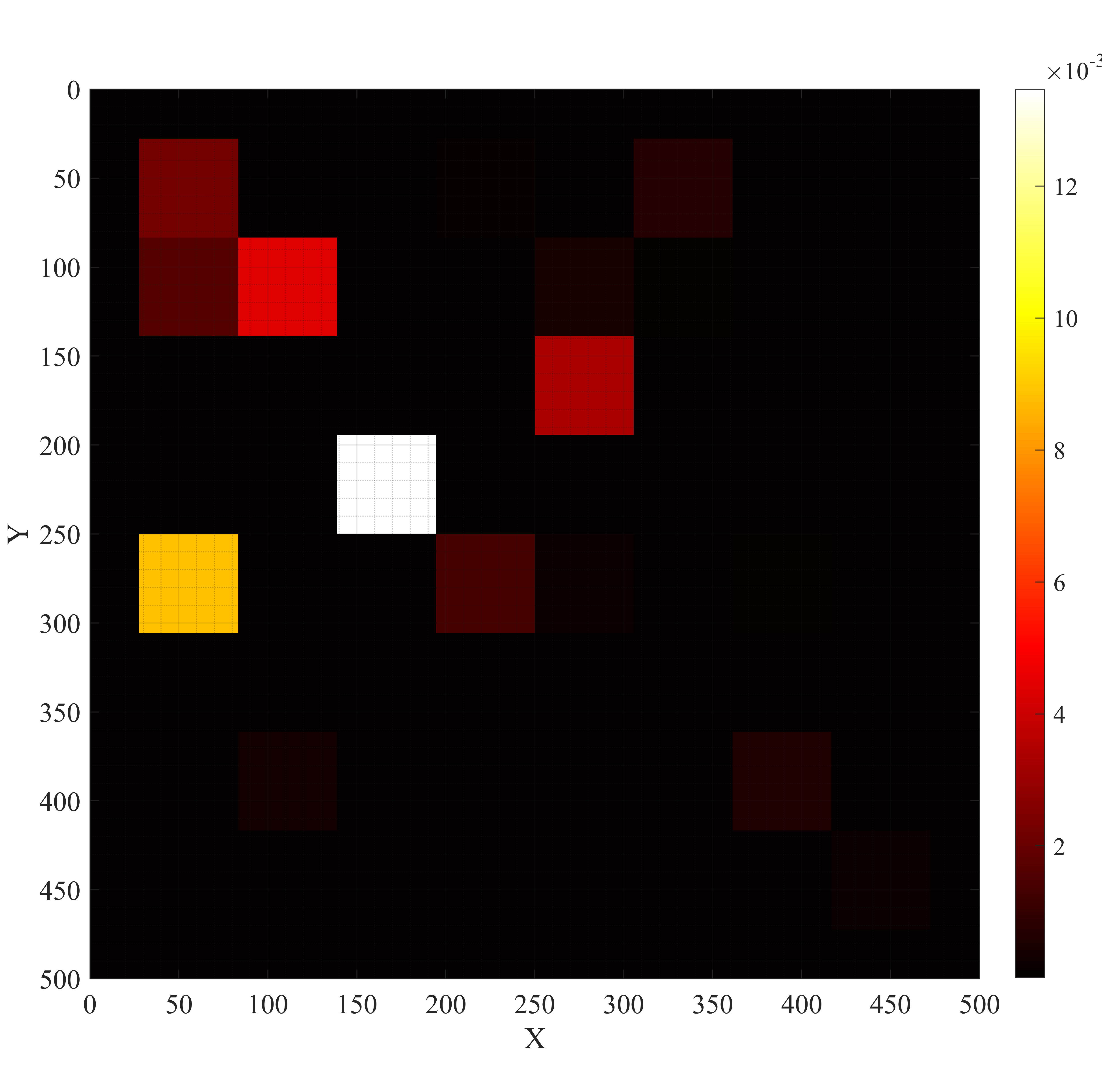}
			%\label{Fig4_a:hswish}
		}	
	\end{minipage}
    \begin{minipage}[t]{0.18\textwidth}
		\centering
		{
			\includegraphics[width=\textwidth]{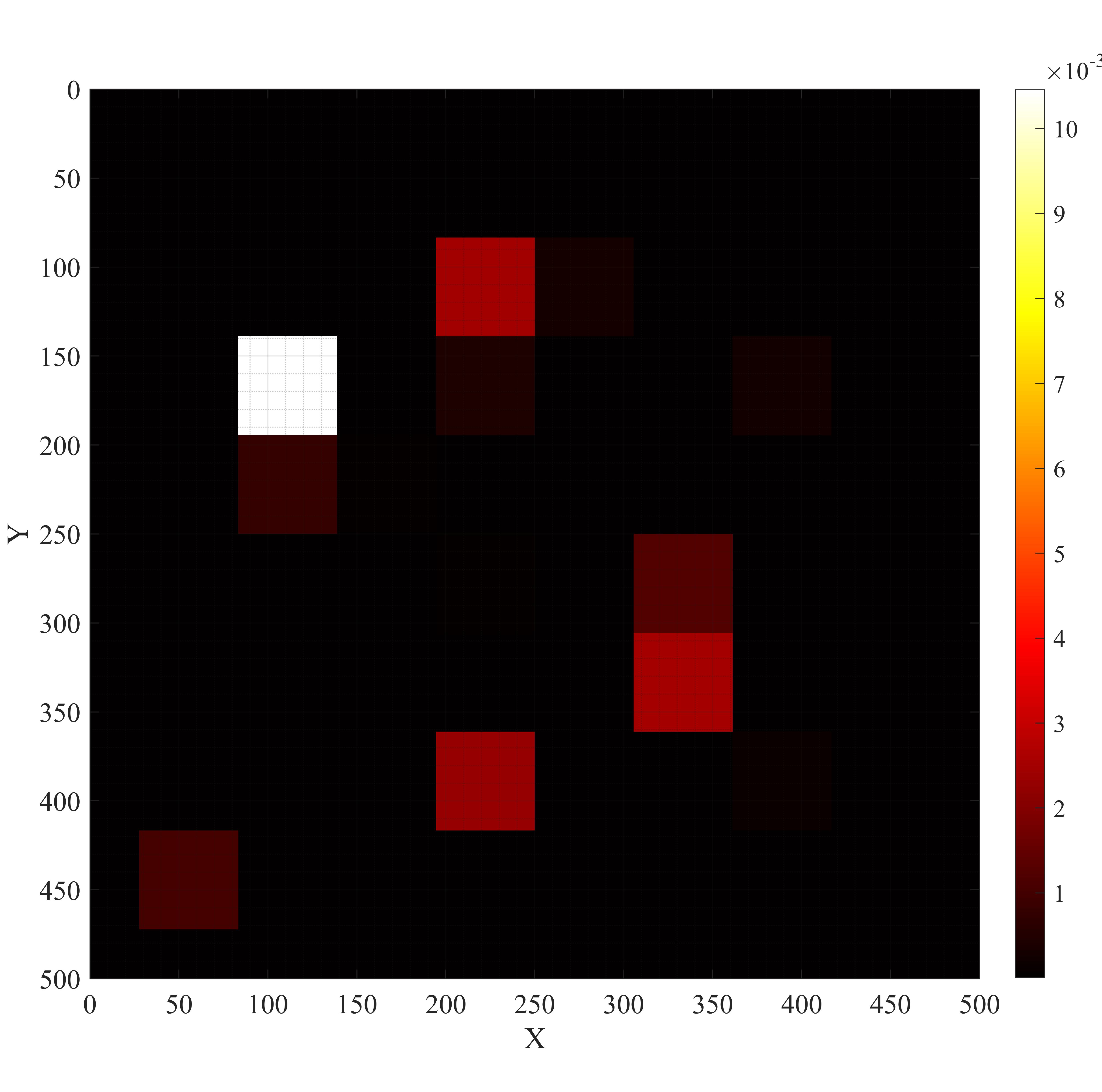}
			%\label{Fig4_a:hswish}
		}	
	\end{minipage}
    \begin{minipage}[t]{0.18\textwidth}
		\centering
		{
			\includegraphics[width=\textwidth]{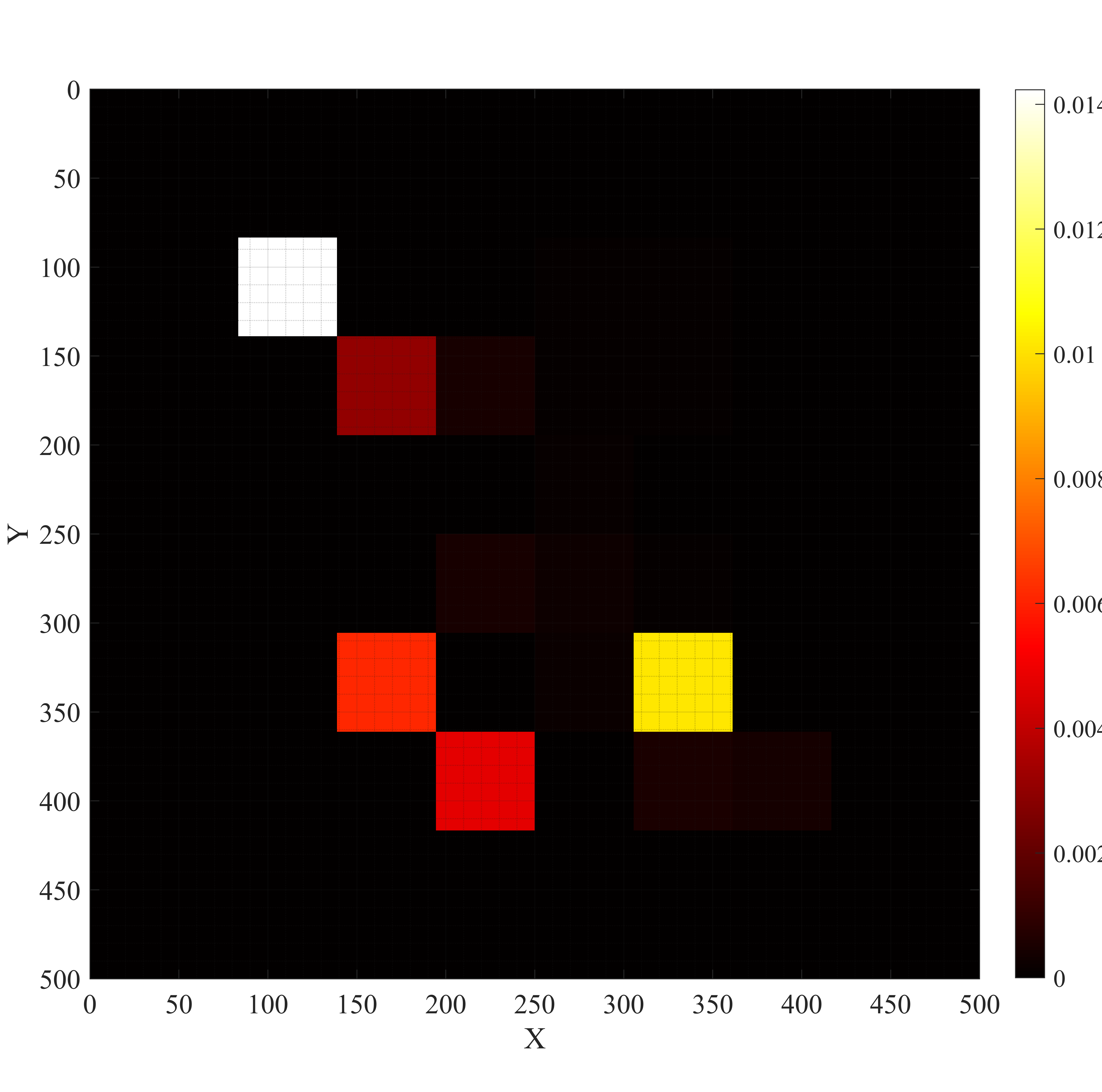}
			%\label{Fig4_a:hswish}
		}	
	\end{minipage}
    \begin{minipage}[t]{0.18\textwidth}
		\centering
		{
			\includegraphics[width=\textwidth]{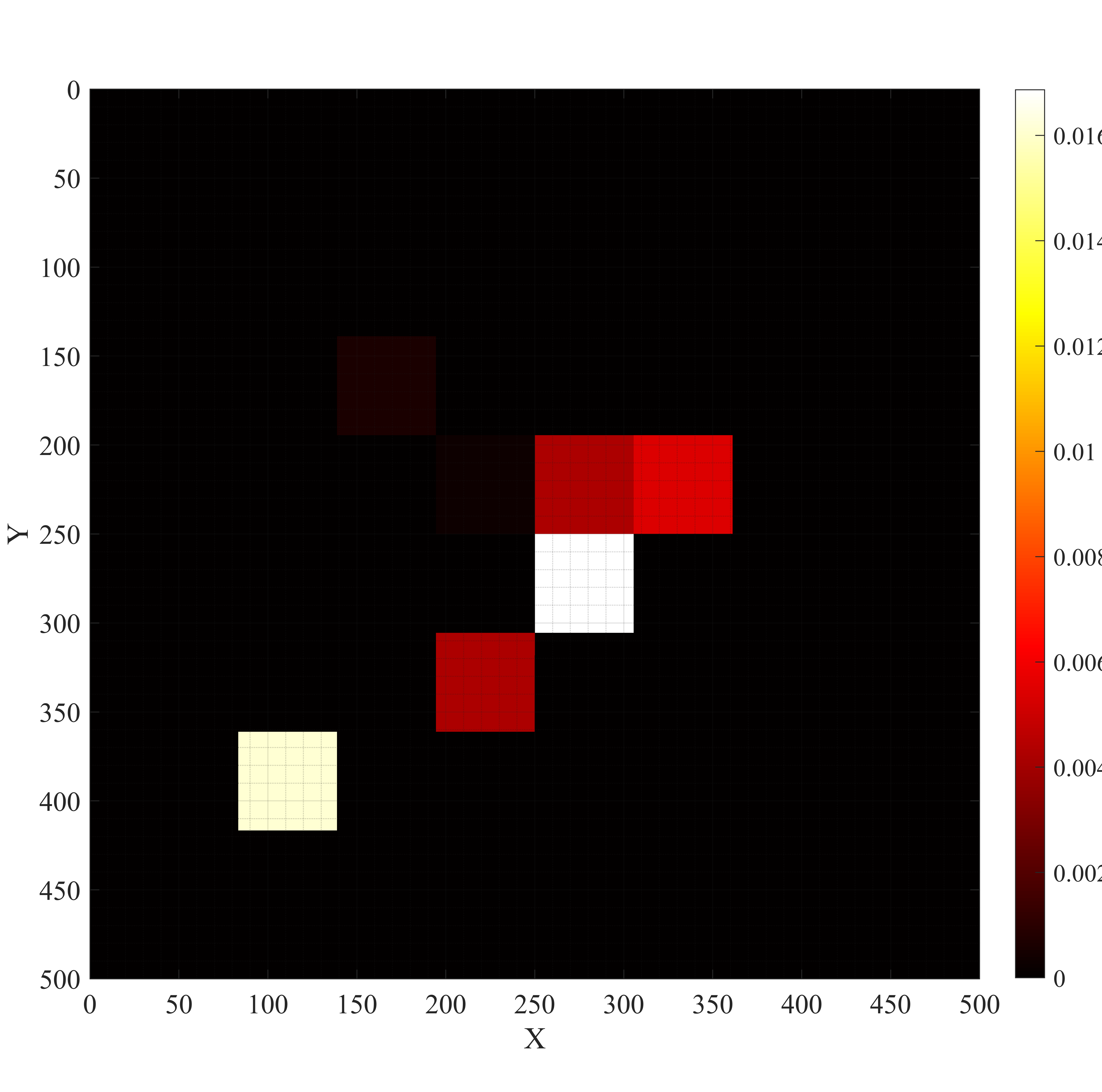}
			%\label{Fig4_a:hswish}
		}	
	\end{minipage} 

    \begin{minipage}[t]{0.18\textwidth}
		\centering
		\subfigure[{
    Step = 0.}]{
			\includegraphics[width=\textwidth]{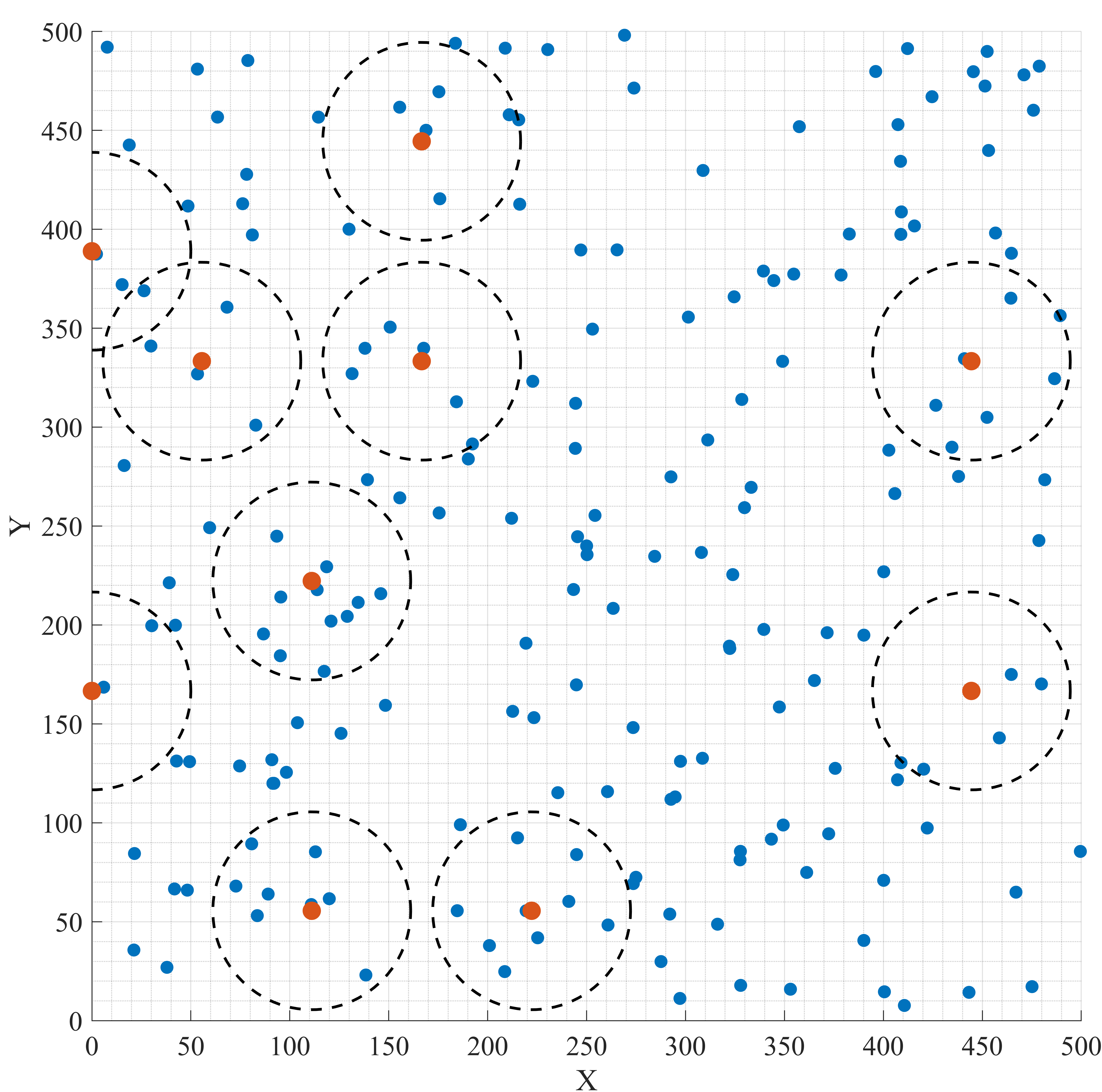}
			%\label{Fig4_a:hswish}
		}	
	\end{minipage}
	\begin{minipage}[t]{0.18\textwidth}
		\centering
		\subfigure[{
    Step = 250.}]{
			\includegraphics[width=\textwidth]{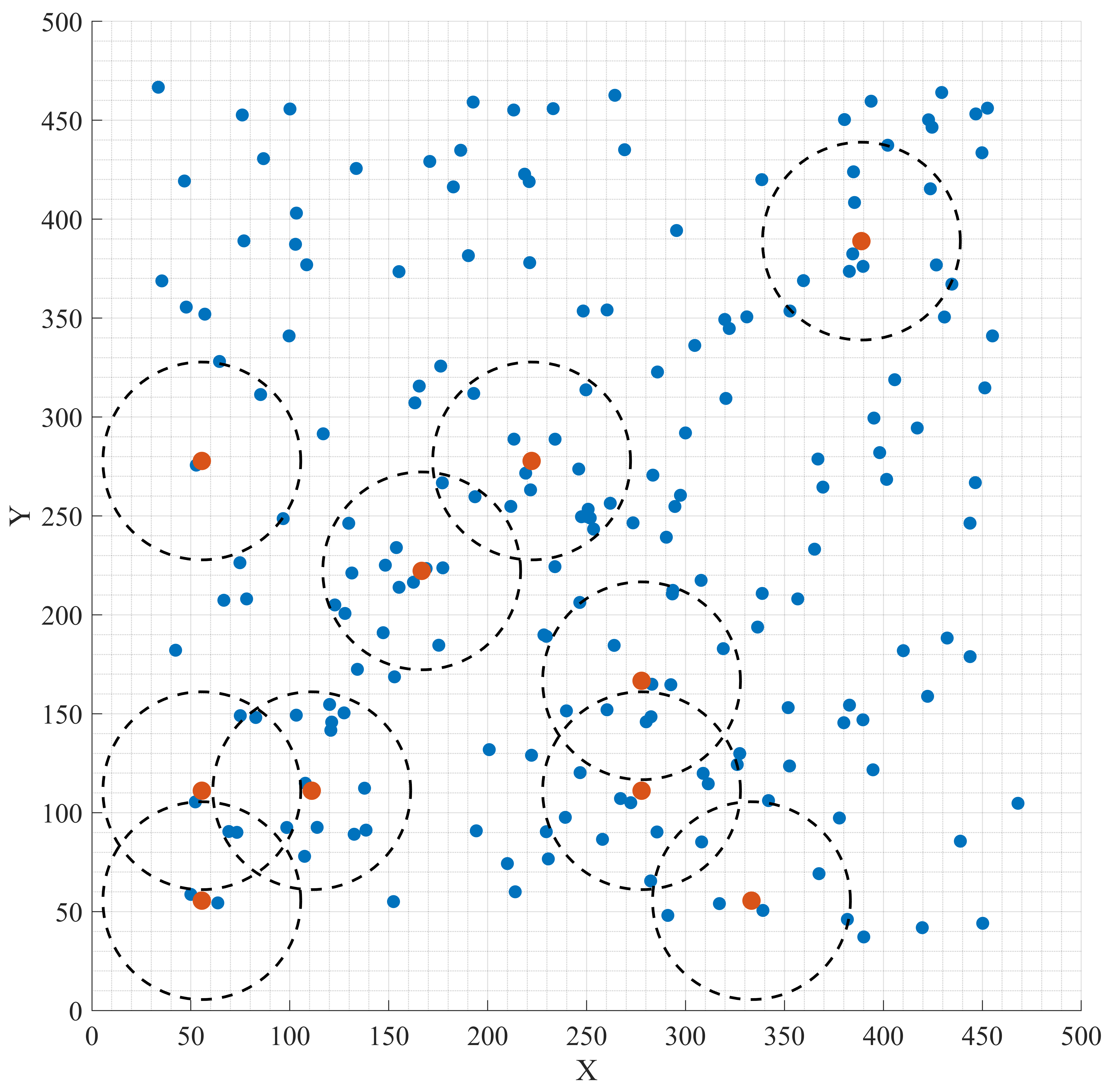}
			%\label{Fig4_a:hswish}
		}	
	\end{minipage}
    \begin{minipage}[t]{0.18\textwidth}
		\centering
		\subfigure[{
    Step = 500.}]{
			\includegraphics[width=\textwidth]{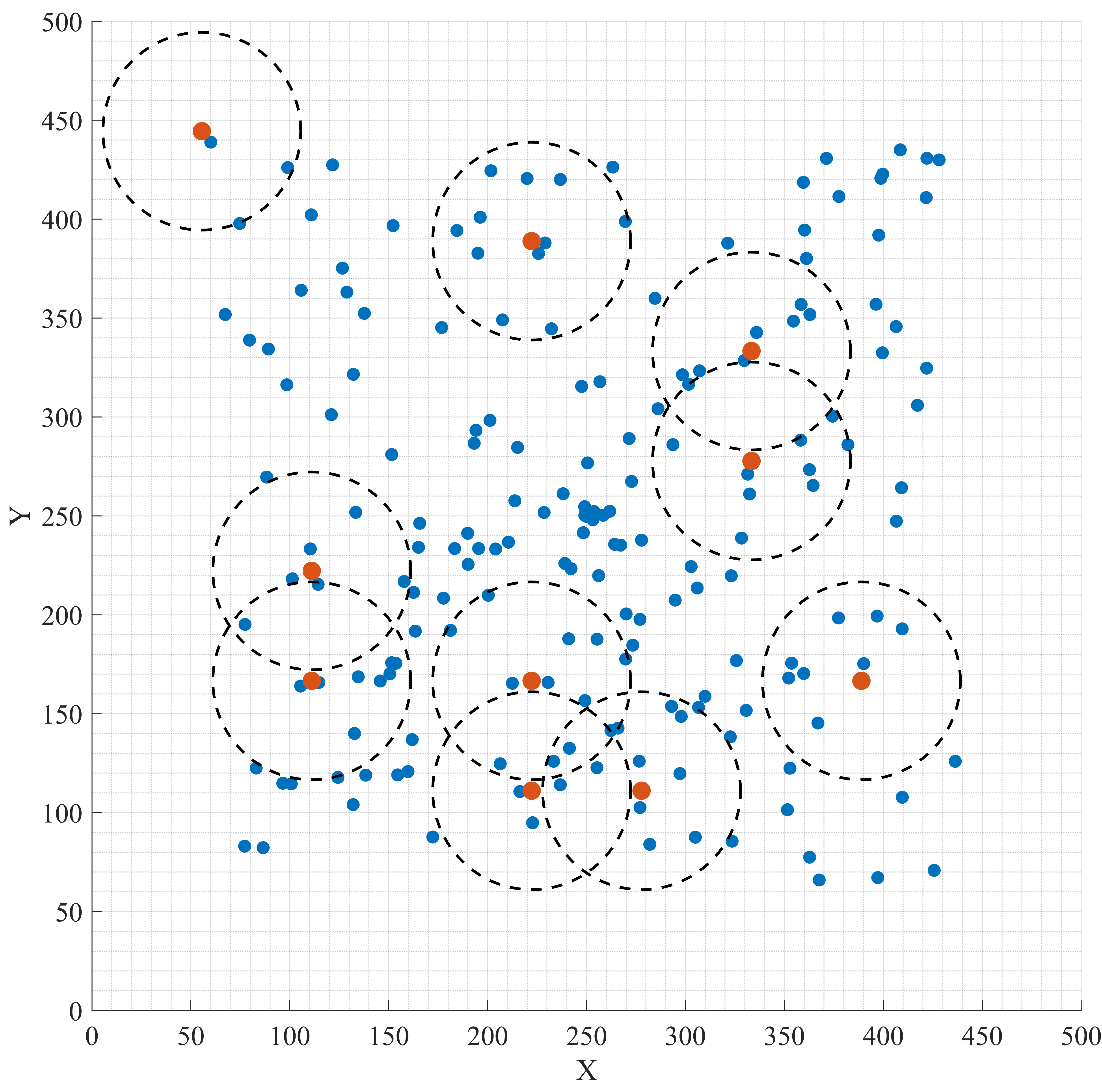}
			%\label{Fig4_a:hswish}
		}	
	\end{minipage}
    \begin{minipage}[t]{0.18\textwidth}
		\centering
		\subfigure[{Step = 750.
    }]{
			\includegraphics[width=\textwidth]{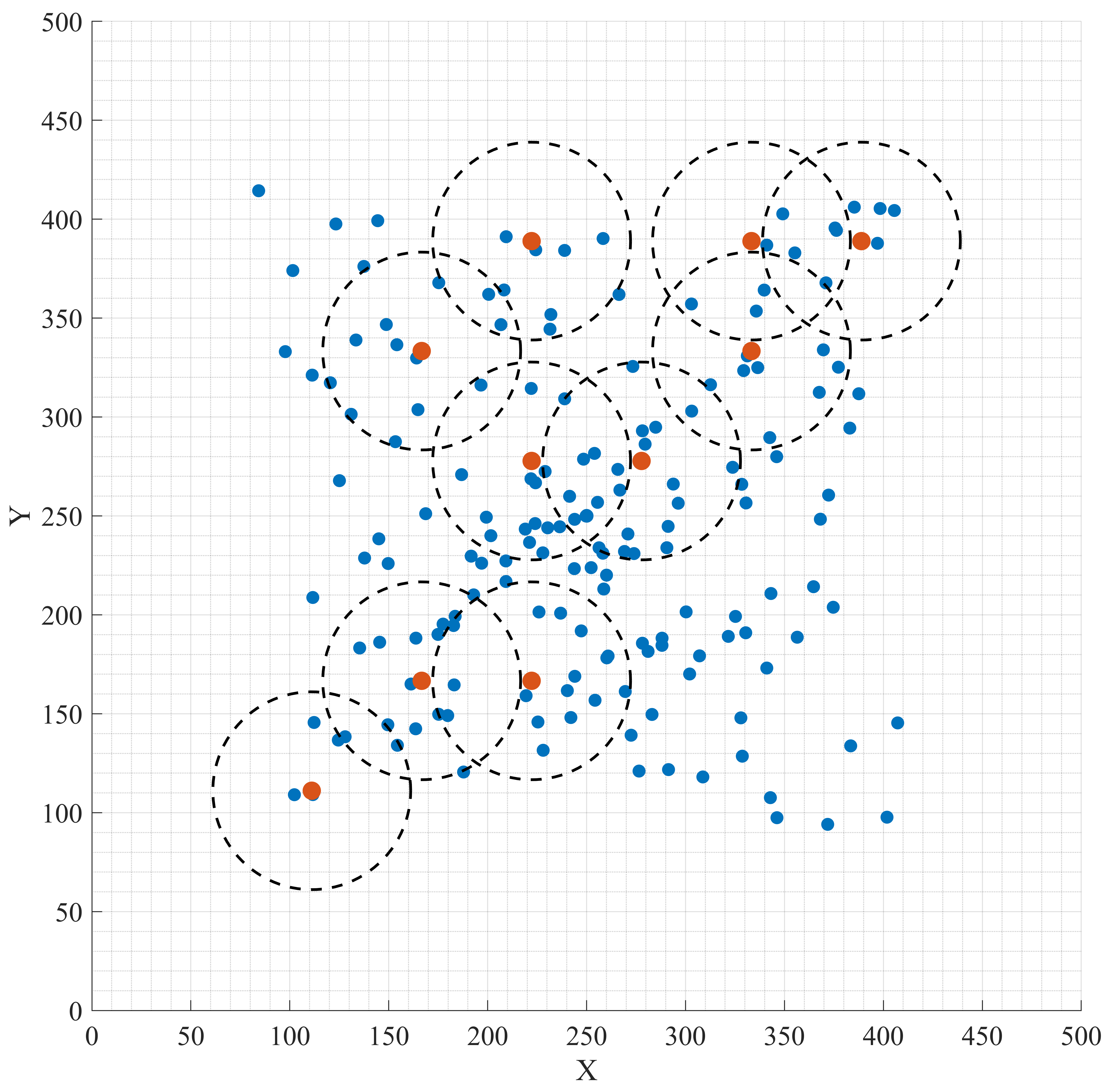}
			%\label{Fig4_a:hswish}
		}	
	\end{minipage}
    \begin{minipage}[t]{0.18\textwidth}
		\centering
		\subfigure[{
    Step = 1000.}]{
			\includegraphics[width=\textwidth]{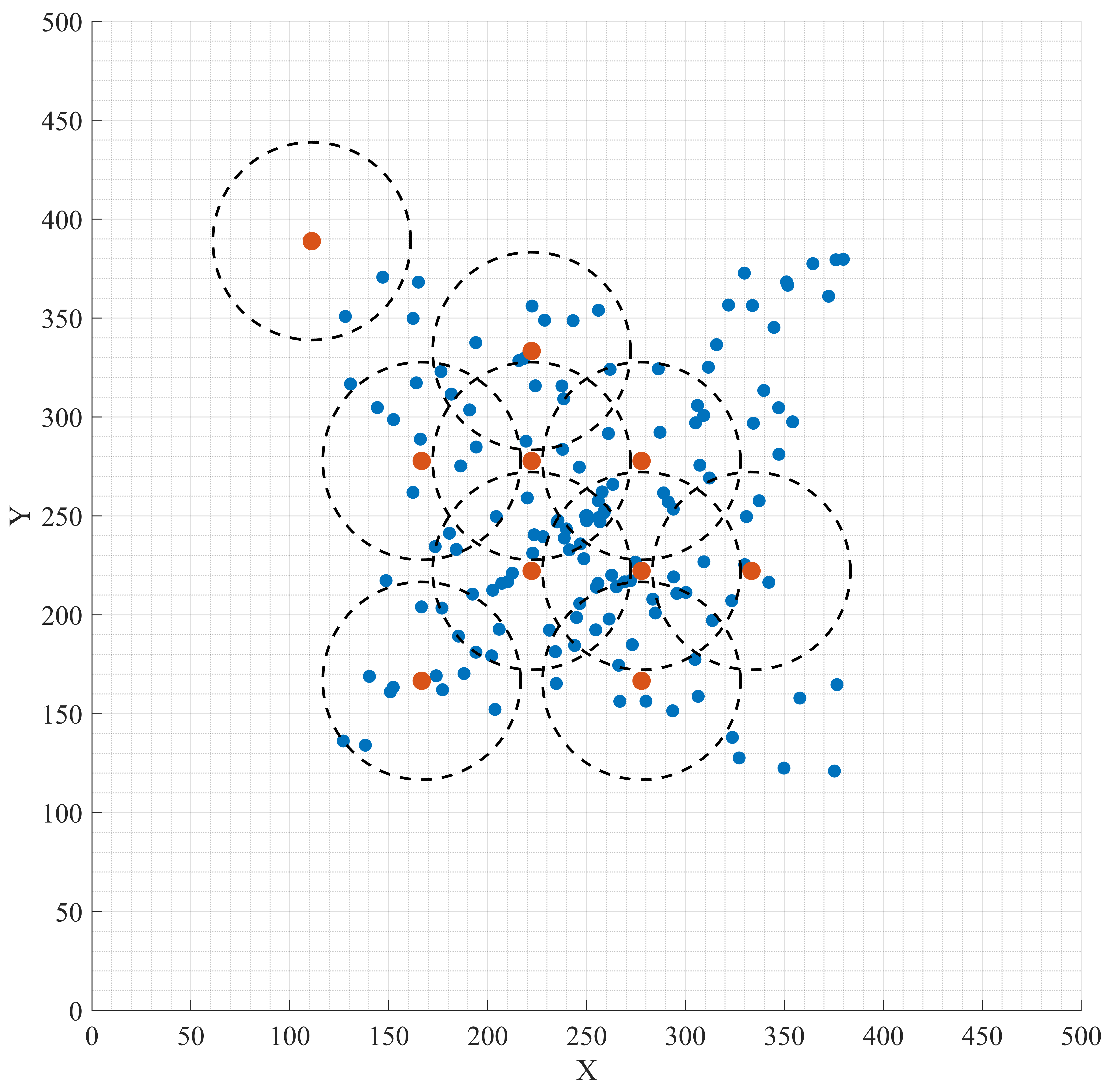}
			%\label{Fig4_a:hswish}
		}	
	\end{minipage}

	\caption{
    {Evaluation for the trained AJ-MAPPO with HDT against a dynamic communication environment and shifting user intents.  
Row 1: plots the trajectories of the AAV (red solid circle) and the user (blue solid circle) across time steps.  
Row 2: converts these paths into a heat-map that quantifies user-density distribution.
Row 3: overlays the AAV’s service range (black hollow circle) on the trajectories, clarifying coverage dynamics.}
\label{fig_all}
    }
\end{figure*}

\subsection{Analysis of HDT and DA-MAPPO}
Fig.~\ref{p_p} presents the training trends of three algorithms: LSTM, Transformer, and HDT. 
The HDT algorithm demonstrates a rapid ascent in accuracy from the outset. Notably, it achieves a remarkable 0.8 prediction accuracy within just 350 episodes. While its accuracy fluctuates slightly in later stages, it consistently stabilizes around 0.75, indicating strong overall performance and reliability.
The Transformer follows a similar upward trajectory but at a more moderate pace. It reaches an accuracy of 0.75 after 450 episodes. Despite not matching HDT's initial speed, it shows steady improvement and stabilizes around 0.75.
In contrast, the LSTM exhibits a more gradual increase in accuracy. It takes approximately 750 episodes to reach an accuracy of 0.6.

Fig.~\ref{drl_p} depicts the training trends of four network optimization in the context of AAV-assisted IoT for intent prediction within IBN. 
DQN reaches a reward of approximately 0.3 at 200 episodes and gradually climbs to around 0.5 by the 1000th episode. 
DDQN demonstrates a notable improvement over DQN, particularly in the initial training phases.
PPO reaches a reward of approximately 0.5 at 150 episodes and continues to increase gradually, stabilizing around 0.7 by the 1000th episode.
DA-MAPPO attains a reward of approximately 0.6 at 100 episodes and continues to rise steadily, reaching around 0.8 by the 1000th episode. 
The numerical results indicate that DA-MAPPO achieves a 33.3\% higher reward (0.8) compared to PPO (0.6) and a 60\% higher reward compared to DDQN (0.5) at the 1000th episode.

Fig.~\ref{drl_f} presents the performance of four decision-making frameworks—DQN, DDQN, PPO, and DA-MAPPO with HDT—in predicting user intent within IBN using AAV-assisted IoT.
DQN exhibits a gradual increase in reward, though it experiences significant volatility throughout the training process. 
DDQN attains a reward of about 0.4 at 150 move steps and continues to rise steadily, reaching approximately 0.55 by the end of 1000 move steps. 
PPO reaches a reward of approximately 0.5 at 100 move steps and continues to increase gradually, stabilizing around 0.6 by the 1000th move step.
DA-MAPPO with HDT outperforms the other frameworks across the entire training process. It achieves a reward of approximately 0.6 at 100 move steps and continues to rise steadily, reaching around 0.7 by the 1000th move step.
The numerical results indicate that DA-MAPPO with HDT achieves a 16.7\% higher reward (0.7) compared to PPO (0.6) and a 27.3\% higher reward compared to DDQN (0.55) at the 1000th move step.

Fig.~\ref{fig_all} offers a comprehensive evaluation of the trained DA-MAPPO with HDT in handling dynamic communication environments and shifting user intents in AAV-assisted IoT. Row 1 clearly shows the trajectories of the AAV (red solid circle) and the user (blue solid circle) at different time steps. At step 0, the AAV and user are at their initial positions. As time progresses to steps 250, 500, 750, and 1000, their positions change dynamically. Row 2 converts these trajectories into heat maps, effectively quantifying the user-density distribution. The heat maps reveal that the user density is concentrated in certain areas, with intensity varying across different steps. Row 3 overlays the AAV's service range (black hollow circle) on the trajectories, providing a clear visualization of coverage dynamics. It is evident that the AAV's service range expands and shifts over time to better cover the user's movement path. The figure demonstrates the effectiveness of DA-MAPPO with HDT in adapting to dynamic changes, maintaining efficient coverage, and improving the accuracy of user intent prediction in AAV-assisted IoT scenarios.

\section{Conclusion}
In this paper, we presented an intent-driven framework that neutralized user-side ambiguity through an implicit intent model, substituted costly matrix-attention operations with HDT, and decoupled intent-response and trajectory-policy sampling via DA-MAPPO. Evaluated on real IoT traces, the system delivered sub-100 ms intent translation and near-optimal network configurations within milliwatt-level edge budgets, surpassing existing baselines.

%\section{ACKNOWLEDGMENT}
% This work was supported by the Key Scientific Research Project of Henan Provincial Higher Education under Grant No. 25A520018.We thank Qwen for assistance with language polishing. 

\bibliographystyle{IEEEtran}
\bibliography{jyz.bib}

@article{pang2020survey,
  title={A survey on intent-driven networks},
  author={Pang, Lei and Yang, Chungang and Chen, Danyang and Song, Yanbo and Guizani, Mohsen},
  journal={Ieee Access},
  volume={8},
  pages={22862--22873},
  year={January, 2020},
  publisher={IEEE}
}

@inproceedings{zeydan2020recent,
  title={Recent advances in intent-based networking: A survey},
  author={Zeydan, Engin and Turk, Yekta},
  booktitle={2020 IEEE 91st vehicular technology conference (VTC2020-Spring)},
  pages={1--5},
  year={May, 2020},
  organization={IEEE}
}

@inproceedings{han2016intent,
  title={An intent-based network virtualization platform for SDN},
  author={Han, Yoonseon and Li, Jian and Hoang, Doan and Yoo, Jae-Hyoung and Hong, James Won-Ki},
  booktitle={2016 12th International Conference on Network and Service Management (CNSM)},
  pages={353--358},
  year={October, 2016},
  organization={IEEE}
}

@article{ouyang2022brief,
  title={A brief survey and implementation on refinement for intent-driven networking},
  author={Ouyang, Ying and Yang, Chungang and Song, Yanbo and Mi, Xinru and Guizani, Mohsen},
  journal={IEEE Network},
  volume={35},
  number={6},
  pages={75--83},
  year={January, 2022},
  publisher={IEEE}
}

@article{yang2023smart,
  title={Smart intent-driven network management},
  author={Yang, Chungang and Mi, Xinru and Ouyang, Ying and Dong, Ru and Guo, Junjie and Guizani, Mohsen},
  journal={IEEE Communications Magazine},
  volume={61},
  number={1},
  pages={106--112},
  year={January, 2023},
  publisher={IEEE}
}

@inproceedings{elkhatib2017charting,
  title={Charting an intent driven network},
  author={Elkhatib, Yehia and Coulson, Geoff and Tyson, Gareth},
  booktitle={2017 13th International Conference on Network and Service Management (CNSM)},
  pages={1--5},
  year={January, 2017},
  organization={IEEE}
}

@article{kiran2018enabling,
  title={Enabling intent to configure scientific networks for high performance demands},
  author={Kiran, Mariam and Pouyoul, Eric and Mercian, Anu and Tierney, Brian and Guok, Chin and Monga, Inder},
  journal={Future Generation Computer Systems},
  volume={79},
  pages={205--214},
  year={February, 2018},
  publisher={Elsevier}
}

@inproceedings{collet2022lossleap,
  title={Lossleap: Learning to predict for intent-based networking},
  author={Collet, Alan and Banchs, Albert and Fiore, Marco},
  booktitle={IEEE INFOCOM 2022-IEEE Conference on Computer Communications},
  pages={2138--2147},
  year={May, 2022},
  organization={IEEE}
}

@inproceedings{li2019graph,
  title={Graph intention network for click-through rate prediction in sponsored search},
  author={Li, Feng and Chen, Zhenrui and Wang, Pengjie and Ren, Yi and Zhang, Di and Zhu, Xiaoyu},
  booktitle={Proceedings of the 42nd international ACM SIGIR conference on research and development in information retrieval},
  pages={961--964},
  year={July, 2019}
}

@article{zyner2019naturalistic,
  title={Naturalistic driver intention and path prediction using recurrent neural networks},
  author={Zyner, Alex and Worrall, Stewart and Nebot, Eduardo},
  journal={IEEE transactions on intelligent transportation systems},
  volume={21},
  number={4},
  pages={1584--1594},
  year={May, 2019},
  publisher={IEEE}
}

@article{xiao2023know,
  title={I know your intent: Graph-enhanced intent-aware user device interaction prediction via contrastive learning},
  author={Xiao, Jingyu and Zou, Qingsong and Li, Qing and Zhao, Dan and Li, Kang and Weng, Zixuan and Li, Ruoyu and Jiang, Yong},
  journal={Proceedings of the ACM on Interactive, Mobile, Wearable and Ubiquitous Technologies},
  volume={7},
  number={3},
  pages={1--28},
  year={September, 2023},
  publisher={ACM New York, NY, USA}
}

@inproceedings{guo2020deep,
  title={A deep prediction network for understanding advertiser intent and satisfaction},
  author={Guo, Liyi and Lu, Rui and Zhang, Haoqi and Jin, Junqi and Zheng, Zhenzhe and Wu, Fan and Li, Jin and Xu, Haiyang and Li, Han and Lu, Wenkai and others},
  booktitle={Proceedings of the 29th ACM International Conference on Information \& Knowledge Management},
  pages={2501--2508},
  year={October, 2020}
}

@article{li2023network,
  title={Network topology optimization via deep reinforcement learning},
  author={Li, Zhuoran and Wang, Xing and Pan, Ling and Zhu, Lin and Wang, Zhendong and Feng, Junlan and Deng, Chao and Huang, Longbo},
  journal={IEEE Transactions on Communications},
  volume={71},
  number={5},
  pages={2847--2859},
  year={February, 2023},
  publisher={IEEE}
}

@article{fang2022drl,
  title={A DRL-driven intelligent optimization strategy for resource allocation in cloud-edge-end cooperation environments},
  author={Fang, Chao and Zhang, Tianyi and Huang, Jingjing and Xu, Hang and Hu, Zhaoming and Yang, Yihui and Wang, Zhuwei and Zhou, Zequan and Luo, Xiling},
  journal={Symmetry},
  volume={14},
  number={10},
  pages={2120},
  year={October, 2022},
  publisher={MDPI}
}

@article{he2023routing,
  title={Routing optimization with deep reinforcement learning in knowledge defined networking},
  author={He, Qiang and Wang, Yu and Wang, Xingwei and Xu, Weiqiang and Li, Fuliang and Yang, Kaiqi and Ma, Lianbo},
  journal={IEEE Transactions on Mobile Computing},
  volume={23},
  number={2},
  pages={1444--1455},
  year={January, 2023},
  publisher={IEEE}
}

@inproceedings{wang2020utility,
  title={Utility optimization for resource allocation in edge network slicing using DRL},
  author={Wang, Zhaoying and Wei, Yifei and Yu, F Richard and Han, Zhu},
  booktitle={GLOBECOM 2020-2020 IEEE Global Communications Conference},
  pages={1--6},
  year={December, 2020},
  organization={IEEE}
}

@article{chen2021drl,
  title={A DRL agent for jointly optimizing computation offloading and resource allocation in MEC},
  author={Chen, Juan and Xing, Huanlai and Xiao, Zhiwen and Xu, Lexi and Tao, Tao},
  journal={IEEE Internet of Things Journal},
  volume={8},
  number={24},
  pages={17508--17524},
  year={May, 2021},
  publisher={IEEE}
}

@article{ullah2023optimizing,
  title={Optimizing task offloading and resource allocation in edge-cloud networks: a DRL approach},
  author={Ullah, Ihsan and Lim, Hyun-Kyo and Seok, Yeong-Jun and Han, Youn-Hee},
  journal={Journal of Cloud Computing},
  volume={12},
  number={1},
  pages={112},
  year={July, 2023},
  publisher={Springer}
}

@article{schulman2017proximal,
  title={Proximal policy optimization algorithms},
  author={Schulman, John and Wolski, Filip and Dhariwal, Prafulla and Radford, Alec and Klimov, Oleg},
  journal={arXiv preprint arXiv:1707.06347},
  year={July, 2017}
}

@inproceedings{jeon2022accurate,
  title={Accurate action recommendation for smart home via two-level encoders and commonsense knowledge},
  author={Jeon, Hyunsik and Kim, Jongjin and Yoon, Hoyoung and Lee, Jaeri and Kang, U},
  booktitle={Proceedings of the 31st ACM International Conference on Information \& Knowledge Management},
  pages={832--841},
  year={October, 2022}
}

@inproceedings{van2016deep,
  title={Deep reinforcement learning with double q-learning},
  author={Van Hasselt, Hado and Guez, Arthur and Silver, David},
  booktitle={Proceedings of the AAAI conference on artificial intelligence},
  volume={30},
  number={1},
  year={March, 2016}
}

@incollection{huang2020deep,
  title={Deep Q-networks},
  author={Huang, Yanhua},
  booktitle={Deep reinforcement learning: fundamentals, research and applications},
  pages={135--160},
  year={June, 2020},
  publisher={Springer}
}

@article{wei2022uav,
  title={UAV-assisted data collection for Internet of Things: A survey},
  author={Wei, Zhiqing and Zhu, Mingyue and Zhang, Ning and Wang, Lin and Zou, Yingying and Meng, Zeyang and Wu, Huici and Feng, Zhiyong},
  journal={IEEE Internet of Things Journal},
  volume={9},
  number={17},
  pages={15460--15483},
  year={May, 2022},
  publisher={IEEE}
}

@article{cheng2023ai,
  title={AI for UAV-assisted IoT applications: A comprehensive review},
  author={Cheng, Nan and Wu, Shen and Wang, Xiucheng and Yin, Zhisheng and Li, Changle and Chen, Wen and Chen, Fangjiong},
  journal={IEEE Internet of Things Journal},
  volume={10},
  number={16},
  pages={14438--14461},
  year={May, 2023},
  publisher={IEEE}
}

@article{wei2020intent,
  title={Intent-based networks for 6G: Insights and challenges},
  author={Wei, Yiming and Peng, Mugen and Liu, Yaqiong},
  journal={Digital Communications and Networks},
  volume={6},
  number={3},
  pages={270--280},
  year={August, 2020},
  publisher={Elsevier}
}

@inproceedings{cerroni2017intent,
  title={Intent-based management and orchestration of heterogeneous openflow/IoT SDN domains},
  author={Cerroni, Walter and Buratti, Chiara and Cerboni, Simone and Davoli, Gianluca and Contoli, Chiara and Foresta, Francesco and Callegati, Franco and Verdone, Roberto},
  booktitle={2017 IEEE Conference on Network Softwarization (NetSoft)},
  pages={1--9},
  year={July, 2017},
  organization={IEEE}
}

@article{wu2020framework,
  title={A framework for off-line operation of smart and traditional devices of IoT services},
  author={Wu, Chung-Yen and Huang, Kuo-Hsuan},
  journal={Sensors},
  volume={20},
  number={21},
  pages={6012},
  year={October, 2020},
  publisher={MDPI}
}

@article{attkan2022cyber,
  title={Cyber-physical security for IoT networks: a comprehensive review on traditional, blockchain and artificial intelligence based key-security},
  author={Attkan, Ankit and Ranga, Virender},
  journal={Complex \& Intelligent Systems},
  volume={8},
  number={4},
  pages={3559--3591},
  year={February, 2022},
  publisher={Springer}
}

@article{custodio2024comparing,
  title={Comparing modern and traditional modeling methods for predicting soil moisture in IoT-based irrigation systems},
  author={Cust{\'o}dio, Gilliard and Prati, Ronaldo Cristiano},
  journal={Smart Agricultural Technology},
  volume={7},
  pages={100397},
  year={March, 2024},
  publisher={Elsevier}
}

@article{pawar2024and,
  title={The What, Why, and How of Context Length Extension Techniques in Large Language Models--A Detailed Survey},
  author={Pawar, Saurav and Tonmoy, SM and Zaman, SM and Jain, Vinija and Chadha, Aman and Das, Amitava},
  journal={arXiv preprint arXiv:2401.07872},
  year={January, 2024}
}

@article{jiang2022efficient,
  title={Efficient planning in a compact latent action space},
  author={Jiang, Zhengyao and Zhang, Tianjun and Janner, Michael and Li, Yueying and Rockt{\"a}schel, Tim and Grefenstette, Edward and Tian, Yuandong},
  journal={arXiv preprint arXiv:2208.10291},
  year={January, 2022}
}

@article{banik2023continuous,
  title={A Continuous Action Space Tree search for INverse desiGn (CASTING) framework for materials discovery},
  author={Banik, Suvo and Loefller, Troy and Manna, Sukriti and Chan, Henry and Srinivasan, Srilok and Darancet, Pierre and Hexemer, Alexander and Sankaranarayanan, Subramanian KRS},
  journal={npj Computational Materials},
  volume={9},
  number={1},
  pages={177},
  year={September, 2023},
  publisher={Nature Publishing Group UK London}
}

@inproceedings{purohit2015intent,
  title={Intent classification of short-text on social media},
  author={Purohit, Hemant and Dong, Guozhu and Shalin, Valerie and Thirunarayan, Krishnaprasad and Sheth, Amit},
  booktitle={2015 ieee international conference on smart city/socialcom/sustaincom (smartcity)},
  pages={222--228},
  year={December, 2015},
  organization={IEEE}
}

@article{graves2012long,
  title={Long short-term memory},
  author={Graves, Alex},
  journal={Supervised sequence labelling with recurrent neural networks},
  pages={37--45},
  year={January, 2012},
  publisher={Springer}
}

@article{han2021transformer,
  title={Transformer in transformer},
  author={Han, Kai and Xiao, An and Wu, Enhua and Guo, Jianyuan and Xu, Chunjing and Wang, Yunhe},
  journal={Advances in neural information processing systems},
  volume={34},
  pages={15908--15919},
  year={May, 2021}
}

@article{mcdonald1995generalization,
  title={A generalization of the beta distribution with applications},
  author={McDonald, James B and Xu, Yexiao J},
  journal={Journal of Econometrics},
  volume={66},
  number={1-2},
  pages={133--152},
  year={March, 1995},
  publisher={Elsevier}
}

@article{li2020first,
  title={First train timetabling for urban rail transit networks with maximum passenger transfer satisfaction},
  author={Li, Xuan and Yamamoto, Toshiyuki and Yan, Tao and Lu, Lili and Ye, Xiaofei},
  journal={Sustainability},
  volume={12},
  number={10},
  pages={4166},
  year={May, 2020},
  publisher={MDPI}
}

@article{zhou2016mobile,
  title={Mobile device-to-device video distribution: Theory and application},
  author={Zhou, Liang},
  journal={ACM Transactions on Multimedia Computing, Communications, and Applications (TOMM)},
  volume={12},
  number={3},
  pages={1--23},
  year={March, 2016},
  publisher={ACM New York, NY, USA}
}

@article{treuille2006continuum,
  title={Continuum crowds},
  author={Treuille, Adrien and Cooper, Seth and Popovi{\'c}, Zoran},
  journal={ACM transactions on graphics (TOG)},
  volume={25},
  number={3},
  pages={1160--1168},
  year={July, 2006},
  publisher={ACM New York, NY, USA}
}

@article{almasan2022deep,
  title={Deep reinforcement learning meets graph neural networks: Exploring a routing optimization use case},
  author={Almasan, Paul and Su{\'a}rez-Varela, Jos{\'e} and Rusek, Krzysztof and Barlet-Ros, Pere and Cabellos-Aparicio, Albert},
  journal={Computer Communications},
  volume={196},
  pages={184--194},
  year={December, 2022},
  publisher={Elsevier}
}

@article{wang2021incorporating,
  title={Incorporating distributed DRL into storage resource optimization of space-air-ground integrated wireless communication network},
  author={Wang, Chao and Liu, Lei and Jiang, Chunxiao and Wang, Shangguang and Zhang, Peiying and Shen, Shigen},
  journal={IEEE Journal of Selected Topics in Signal Processing},
  volume={16},
  number={3},
  pages={434--446},
  year={December, 2021},
  publisher={IEEE}
}

% \newpage

% \begin{IEEEbiography}[{\includegraphics[width=1in,height=1.25in, clip, keepaspectratio]{Zhengwei Xu.jpg}}]{Zhengwei Xu} received the Ph.D. degree in Computer Science and Technology from Hohai University, Nanjing, China, in 2023. He is currently a Lecturer with the College of Computer and Information Engineering, Henan Normal University, Henan, China. His current research interests include specific emitter identification, industrial Internet of Things, and deep learning.
% \end{IEEEbiography}

% \begin{IEEEbiography}[{\includegraphics[width=1in,height=1.25in, clip, keepaspectratio]{Shaopeng Lu.png}}]{Shaopeng Lu} received the B.S. degree in software engineering from Zhengzhou Normal University, Zhengzhou, China, in 2023. He is currently pursuing a master's degree at Henan Normal University. 
% His research focuses on specific emitter identification, broad learning systems, and spiking neural networks.
% \end{IEEEbiography}

% \begin{IEEEbiography}[{\includegraphics[width=1in,height=1.25in, clip, keepaspectratio]{Zhihao Ou.jpg}}]{Zhihao Ou} received the B.S. andPh.D.degrees in computer science from NanjingUniversity, Nanjing, China, in 2009 and 2018, respectively.Currently, he is an Associate Professorwith the College of Computer Science and SoftwareEngineering, Hohai University. His research interests include wireless networks, edge computing, anddistributed machine learning 
% \end{IEEEbiography}

\vfill

\end{document}